\documentclass[lettersize,journal]{IEEEtran}
\usepackage{amsmath,amsfonts}
\usepackage{algorithmic}
\usepackage{algorithm}
\usepackage{array}
\usepackage[caption=false,font=normalsize,labelfont=sf,textfont=sf]{subfig}
\usepackage{textcomp}
\usepackage{stfloats}
\usepackage{siunitx} 
\usepackage{url}
\usepackage{verbatim}
\usepackage{graphicx}
\usepackage{cite}
\usepackage{multirow}
\usepackage{amsfonts}
\usepackage{threeparttable}
\usepackage{color}
\usepackage{xcolor}
\usepackage{colortbl}
\usepackage{makecell}
\usepackage{bm}
\usepackage{epstopdf}
\usepackage{url}
\usepackage{booktabs}
\newsavebox{\mybox}

\newcommand{\magblue}[1]{{\color{magenta} \textbf{#1}}}
\begin{document}

\title{FAS-UNet: A Novel FAS-driven Unet to Learn Variational Image Segmentation\thanks{This work was supported by the National Natural Science Foundation
of China (NSFC) under Grants 11971414, 11771369, also partly by grants from Natural
Science Foundation of Hunan Province under Grants 2018JJ2375, 2018XK2304, and 2018WK4006. (Corresponding author: Jianping Zhang).}}

\author{Hui Zhu\thanks{H. Zhu is with the School of Mathematics and Computational Science, Xiangtan University, and Key Laboratory of Intelligent Computing \& Information Processing of Ministry of Education (201931000089@smail.xtu.edu.cn.}, Shi Shu\thanks{S. Shu is with the School of Mathematics and Computational Science, Xiangtan University, and Hunan Key Laboratory for Computation and Simulation in Science and Engineering, Xiangtan, 411105, China (shushi@xtu.edu.cn).}, and Jianping Zhang\thanks{J. Zhang is with the School of Mathematics and Computational Science, Xiangtan University, and Hunan National Applied Mathematics Center (jpzhang@xtu.edu.cn).}}


\markboth{
}%
{Shell \MakeLowercase{\textit{et al.}}: A Sample Article Using IEEEtran.cls for IEEE Journals}

\IEEEpubid{
}

\maketitle

\begin{abstract}
Solving variational image segmentation problems with
hidden physics is often expensive and requires different algorithms and manually tunes model parameter.
The deep learning methods based on the U-Net structure have obtained outstanding performances in many different medical image segmentation tasks, but designing such networks requires a lot of parameters and training data, not always available for practical problems.
In this paper, inspired by traditional multi-phase convexity Mumford-Shah variational model and full approximation scheme (FAS) solving the nonlinear systems, we propose a novel variational-model-informed network (denoted as FAS-Unet) that exploits the model and algorithm priors to extract
the multi-scale features. The proposed model-informed network integrates image data and mathematical models, and implements them through learning a few convolution kernels.  Based on the variational theory and FAS algorithm, we first design a feature extraction  sub-network (FAS-Solution module) to solve the model-driven nonlinear systems, where a skip-connection is employed  to
fuse the multi-scale features. Secondly, we further design a convolution block to fuse the extracted features from the previous stage, resulting in the final segmentation possibility.
Experimental results on three different medical image segmentation tasks show that the proposed FAS-Unet is very competitive with other state-of-the-art methods in qualitative, quantitative and model complexity evaluations.
Moreover, it may also be possible to train specialized network
architectures that automatically satisfy some of the mathematical and physical laws in other image problems for better accuracy, faster
training and improved generalization. \magblue{The code is available at \url{https://github.com/zhuhui100/FASUNet}}.
\end{abstract}


\section{Introduction}

Image segmentation is one of the most important problems in computer vision and~also is a difficult problem in the medical imaging community~\cite{minaee_ImageSegmentation_2021a,boveiri_MedicalImage_2020,cai_ReviewApplication_2020}.
 It has been widely used in many medical image processing fields such as the identification of cardiovascular
 {diseases} \cite{chen_DeepLearning_2020}, the~measurement of bone and tissue~\cite{litjens_SurveyDeep_2017}, and~the extraction of suspicious lesions to aid radiologists.
 Therefore, image segmentation has a
vital role in promoting medical image analysis and applications as a powerful image processing tool~\cite{litjens_SurveyDeep_2017,miotto_DeepLearning_2018}.

Deep learning (DL) has achieved great success in the field of medical image segmentation~\cite{fu_DeepLearning_2020,liu_ReviewDeepLearningBased_2021,litjens_SurveyDeep_2017}. One of the most important reasons is that the convolutional neural networks (CNNs) can effectively extract image features. Therefore, much work at present involves design a network architecture with strong feature extraction ability, and~many well-known CNN architectures have been proposed such as UNet~\cite{UNet}, V-Net~\cite{milletari_VNetFully_2016a}, UNet++ \cite{zhou2018nest}, 3D UNet~\cite{cicek_3DUNet_2016a}, Y-Net~\cite{SMehta2018}, Res-UNet~\cite{xiao_WeightedResUNet_2018}, KiU-Net~\cite{valanarasu_KiUNetAccurate_2020}, DenseUNet~\cite{li_HDenseUNetHybrid_2018a}, and~nnU-Net~\cite{IsenseeF_2021}. More and more studies based on data-driven methods have been reported for medical image segmentation. Although~UNet and its variants have achieved considerably impressive performance in many medical image segmentation datasets, they still suffer two limitations. One is that most of researchers have introduced more parameters to improve the performance of medical image segmentation, but have tended to ignore the technical branch of the model's memory and computational overhead, which makes it difficult to popularize the algorithm to industry applications~\cite{He2019ODEInspiredND}. The~other disadvantage is that these variants only design many suitable architectures through the researcher's experience or experiments, but do not focus on the mathematical theoretical guidance of network architectures such as explainability, generalizability, etc., which limits the application of these models and the improvement of task-driven medical image segmentation methods~\cite{pmlr-v80-lu18d,canet2020}.

Recently, many works on image recognition and image reconstruction have been focusing on the interpretability of the network architecture. Inspired by some mathematical viewpoints, many related unroll networks have been designed and successfully applied. He~et~al.~\cite{he2016deep} proposed the deep residual learning framework, which utilizes an identity map to facilitate training; it is well known that it is very similar to the iterative method solving ordinary {differential} equations (ODEs) and~also achieves promising performance on image recognition. G. Larsson et al employed the fractal idea to design a self-similar FractalNet~\cite{larsson2017fractalnet}, also discovering that its architecture is similar to the Runge--Kutta (RK) scheme in numerical calculations. According to the nature of polynomials, Zhang~et~al. designed PolyNet~\cite{zhang2017polynet} by improving ResNet to strengthen the expressive ability of the network, and~Gomez~et~al.~\cite{gomez2017reversible} proposed RevNet by using some ideas of the dynamic system. Chen~et~al.~\cite{chen2018neural} analyzed the process of solving ODEs, then proposed Neural ODE, which further shows that mathematics and neural networks have a strong relationship. Meanwhile, He et al. designed a network architecture for the super-resolution task based on the forward Euler and RK methods of solving ODEs~\cite{He2019ODEInspiredND} and~achieved good performance. Sun~et~al.~\cite{Yang0LX16} designed ADMM-Net through the alternating direction method to learn an image reconstruction problem. Inspired by a multigrid algorithm for solving inverse problems, He~et~al.~\cite{He2019MgNetAU} proposed a learnable classification network denoted as MgNet to extract image features $\bm{u}$, which uses a few parameters to achieve good performance on the CIFAR datasets. Alt~et~al.~\cite{alt2021connections} analyzed the behavior and mathematical foundations of UNet, and~interpreted them as approximations of continuous nonlinear partial differential equations (PDEs) by using full approximation schemes (FASs). Experimental evaluations showed that the proposed architectures for the denoising and inpainting tasks save half of the trainable parameters and can thus outperform standard ones with the same model~complexity.

Unfortunately, only a few studies based on model-driven techniques have been reported for the segmentation task. In~this paper, we mainly focus on the explainable DL framework combining the advantages of the FAS and UNet for medical image~segmentation.

\begin{figure*}[htbp]
\centering
\includegraphics[width=0.99\textwidth]{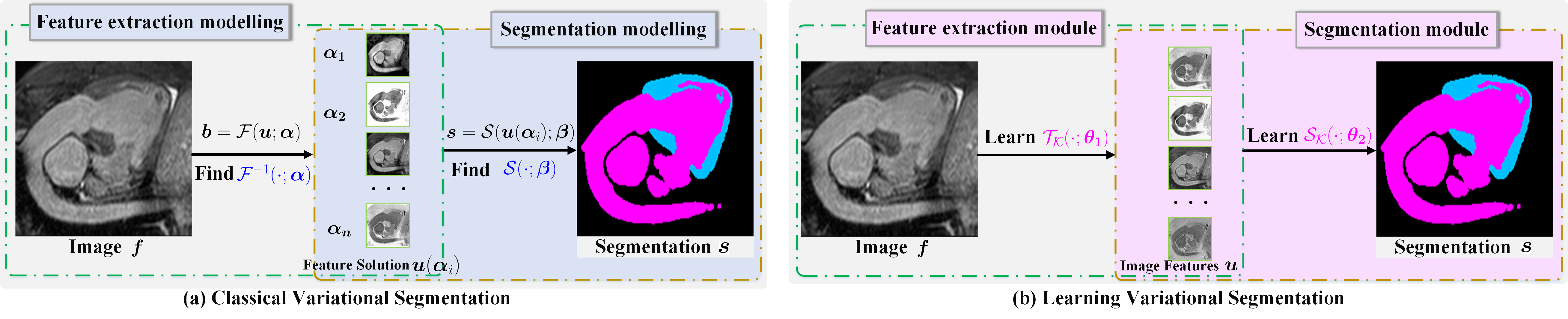}
  \caption{Classical variational image segmentation and model-inspired learning method. (a) The first stage solves the nonlinear differential equations using classical iterative method, and then the second stage thresholds
the smooth solution in the first stage to extract objects. (b) The first stage learns the solution mapping $\mathcal{T}_{\mathcal{K}}(f;\bm{\theta}_1)$ by optimizing the convolution kernel $\bm{\theta_1}$ to extract image features, The second stage learns feature fusion and segmentation thresholding parameter.}
  \label{Image_segmentation}
\end{figure*}

\subsection{Problem} H. Helmholtz proposed that the ill-posed problem of producing reliable perception from fuzzy signals can be solved through the process of ``unconscious inference'' (the \textbf{Helmholtz Hypothesis}) \cite{GHatfield2002}. This theory implies that human vision is incomplete and that details are inferred by the unconscious mind to create a complete image. That is, our perception system can also integrate the fuzzy evidence received from the senses into the situation based on its own environmental~model.

Let $p(\bm{u}|\bm{f};\bm{\alpha})$ be a probabilistic distribution for feature
representations $\bm{u}$ of the source image $\bm{f}$.
The prior probability of $\bm{u}$ can be modeled as the
multivariate normal distribution. In~general, $\bm{u}$ can be extracted from a given image $\bm{f}$ by optimizing the maximum a posteriori (MAP) estimation as 


\begin{equation}\label{eq_feature_1.1}
\arg\max\limits_{\bm{u}} \log p(\bm{u}|\bm{f};\bm{\alpha}),
\end{equation}
where $\bm{\alpha}$ is the environmental parameter in classical ``unconscious inference'' or the inverse problem, and~this problem leads to the nonlinear system defined by
\begin{equation}\label{eq_feature_1a1}
\mathcal{F}(\bm{u};\bm{\alpha})=\bm{b},
\end{equation}
where the nonlinear operator $\mathcal{F}(\cdot;\bm{\alpha})$ is employed to generate the image $\bm{b}$, e.g.,~$\bm{b}=\mathcal{A}^T\bm{f}$ is a deconvoluted image of $\bm{f}$ in the image deblurring problem with a convolution operator~$\mathcal{A}$.

We consider that image segmentation refers to a composite process of feature extraction (\ref{eq_feature_1a}) and feature fusion segmentation.
Here, the~fusing process for feature $\bm{u}$ is defined by
\begin{equation}
\bm{s}=\mathcal{S}(\bm{u};\bm{\beta}),
\label{eq0a}
\end{equation}
where $\mathcal{S}(\cdot;\bm{\beta})$ denotes a fusing segmentation with a fixed conscious parameter $\bm{\beta}$, and $\bm{s}$ is the segmentation results or probability~maps.

Such strongly interpretable segmentation models~\cite{XHCai2013,liu_WeightedVariational_2018,ma_ImageSegmentation_2018} are so general that, depending on the amount of well-predefined sparsity priors of the input image, they have the advantages of theoretical support and strong convergence. The~total flowchart of classical variational segmentation can be summarized as shown in Figure~\ref{Image_segmentation}a. However, they usually require expensive computations, but~also have to face the problems of the selection of suitable regularizers $\phi(\cdot)$ and model parameters $(\bm{\alpha},\bm{\beta})$. Consequently, some reconstructed results are
~unsatisfactory.

It is well known that the solution $\bm{u}$ usually has the multiscale property, so a natural idea is to exploit the multi-layer convolution and multigrid architecture, which can describe multiscale features to learn $\bm{u}$. Based on the above facts, we propose a two-stage segmentation framework for learning feature $\bm{u}$ in Stage 1 and segmentation $\bm{s}$ in Stage 2, which is shown in Figure~\ref{Image_segmentation}b.

\subsection{Contributions} In this work, we focus on analyzing the feature extraction inverse problem (\ref{eq_feature_1a1}) and the feature fusion segmentation (\ref{eq0a}) to design an explainable deep learning network. It is well known that the unrolled iterations of the classical solution algorithm can be considered as the layers of a neural network, so we propose a novel FAS-driven UNet (FAS-UNet), which integrates image data and a multiscale algorithm for solving the nonlinear inverse problem (\ref{eq_feature_2}).
The major differences with our approach are that MgNet is not a U-shaped architecture and is only used for image classification, which leads to the output result not being able to be converted to the
segmentation prediction of the input image. Besides, the~proposed network was inspired by the traditional multiphase convexity Mumford--Shah variational model~\cite{XHCai2013} and FAS algorithm for solving nonlinear systems~\cite{mccormick1987multigrid}, which exploits the model and algorithm priors' information to extract the image features.
Indeed, the~goal of our work is to show that, under~some assumptions about the operators, it is possible to interpret the smoothing operations of the FAS and image geometric extracting operations of the variational model as the layers of a CNN, which in turn, provide fairly specialized network architectures that allow us to solve the standard nonlinear system (\ref{eq_feature_2}) for a specific choice of the parameters~involved.

Our main contributions are summarized as~follows:
\begin{enumerate}
\item[1.]
We propose a novel variational-model-informed two-stage image segmentation network (FAS-UNet), where an explainable and lightweight sub-network for feature extraction is designed by combining the traditional multiphase
convexity Mumford--Shah variational model and FAS algorithm for solving nonlinear systems. To~the best of our knowledge, it is the first unrolled architecture designed based on model and algorithm priors in the image segmentation community.
\item[2.] The proposed model-informed network integrates image data and mathematical models, and it provides a helpful viewpoint for designing the image segmentation network architecture.
\item[3.] The proposed architecture can be trained from additional model information obtained by enforcing some of the mathematical and physical laws for better accuracy, faster training, and improved generalization. Extensive experimental results show that it performs better than the other state-of-the-art methods.
\end{enumerate}

The rest of the paper is organized as follows. The~novel FAS-UNet framework for solving nonlinear inverse problems by analyzing variational segmentation theory and the FAS algorithm is proposed in Section~\ref{sect2}. We show
experimental results in Section~\ref{sect3}. Finally, we conclude this work in Section~\ref{sect4}.

\section{Variational Segmentation via the CNN~Framework}\label{sect2} The goal of image segmentation is to partition a given image $f: \Omega \rightarrow \mathbb{R}$ into $r$ regions $\left\{\Omega_{i}\right\}_{i=1}^{r}$ that contain distinct objects and satisfy $\Omega_{i} \cap \Omega_{j}=\emptyset, j \neq i$, and~$\bigcup_{i=1}^{r} \Omega_{i}=\Omega$, where the image domain $\Omega$ is a bounded and open subset of $\mathbb{R}^{2}$. Assume that $\Gamma=\bigcup \partial \Omega_{i}$ is the union of boundaries of $\Omega_{i},|\Gamma|$, denoting the arc length of curve $\Gamma$.
\subsection{Multiphase Variational Image~Segmentation}
As mentioned, various ways of variational image
segmentation have been proposed. Below,~we review a few of~them.

\subsubsection{Variational Image~Segmentation} The Mumford--Shah (M-S) model is a well-known variational image segmentation
method proposed by Mumford and Shah~\cite{mumford_OptimalApproximations_1989a}, which can be defined as follows:
\begin{equation*}\label{eq_MS_model}
\min\limits_{u,\Gamma} \left\{\tau_1 \int_{\Omega}(f-u)^{2} d \bm{x}+\tau_2 \int_{\Omega-\Gamma}|\nabla u|^{2} d \bm{x}+|\Gamma|\right\},
\end{equation*}
where $\tau_1$ and $\tau_2$ are the weight parameters. The~first term requires that $u: \Omega \rightarrow \mathbb{R}$ approximates $f$, the~second term that $u$ does not vary much on each $\Omega_{i}$, and~the third term that the boundary $\Gamma$ is as short as possible. This shows that $u$ is a piecewise smooth approximation of $f$.

In particular, Chan and Vese considered the special case of the M-S model where the function $u$ is chosen to be a piecewise constant function; thus, the minimization for two-phase segmentation is given as
\begin{equation*}
\begin{split}
\min\limits_{\Gamma,c_1,c_2}\lambda_{1}\int_{\text {inside }(\Gamma)}\left|f-c_{1}\right|^{2} d \bm{x}
+\lambda_{2} \int_{\text {outside }(\Gamma)}\left|f-c_{2}\right|^{2} d \bm{x}+|\Gamma|,
\end{split}
\end{equation*}
where $c_1$ and $c_2$ are the average image intensities inside and outside of boundary $\Gamma$, respectively, and $\lambda_1$ and $\lambda_2$ are the weight~parameters.

Sometimes, the~given image is degraded by noise and problem-related blur operator $\mathcal{A}$. Therefore, Cai~et~al.~\cite{XHCai2013} extended the two-stage image segmentation strategy using a convex variant of the Mumford--Shah model as
\begin{equation}\label{cai-eq}
\min_{u \in \text{W}^{1,2}(\Omega)}\int_{\Omega}\left(\kappa_1(f-\mathcal{A}u)^{2}+\kappa_2|\nabla u|^{2}+|\nabla u|\right) d \bm{x},
\end{equation}
where $\kappa_1$ and $\kappa_2$ are positive parameters, and~the existence and uniqueness of $u$ were analyzed in their~work.

 We assume the image features $\bm{u}=(\bm{u}_1,\dots,\bm{u}_d)^T:\Omega\rightarrow\mathbb{R}^{d}$, where $\bm{u}_{i}:\Omega_i\rightarrow\mathbb{R}$ is a smooth mapping defined on the tissue or lesion $\Omega_{i}$. In~this work, we extend the above model (\ref{cai-eq}) to the multiphase case, which can deal with $d$-phase segmentation (multiple objects), which refers to a two-stage composite process of feature extraction (\ref{eq_feature_1a}) and feature fusion segmentation (\ref{eq0a}).

\subsubsection{Feature~Extraction} The first stage is to extract image features $\bm{u}$ by maximizing a posterior probabilistic distribution (\ref{eq_feature_1a}) for feature representations $\bm{u}$ of a given image $f$ as
\begin{equation}\label{eq_feature_1}
\begin{split}
\arg\max\limits_{\bm{u}} p(\bm{u}|f;\bm{\alpha})&=\arg\max\limits_{\bm{u}}\log p(\bm{u}|f;\bm{\alpha})\\
&= \arg\max\limits_{\bm{u}} \log \frac{p(f|\bm{u};\bm{\alpha})p(\bm{u};\bm{\alpha})}{p(f)}\\
&= \arg\max\limits_{\bm{u}} \log p(f|\bm{u};\bm{\alpha})p(\bm{u};\bm{\alpha}),
\end{split}
\end{equation}
where $\bm{\alpha}$ is the environmental parameter in classical ``unconscious inference'' or the inverse problem. Especially, the~likelihood
probability $p(f|\bm{u};\bm{\alpha})$ and the prior probability $p(\bm{u};\bm{\alpha})$ can be modeled as normal distributions, respectively, denoted by
\begin{equation*}
\begin{split}
&p(f|\bm{u};\bm{\alpha})\propto e^{-\frac{1}{2\sigma^2}\int_\Omega(\mathcal{A}\bm{u}-f)^2d\Omega}=e^{-\gamma\int_\Omega(\mathcal{A}\bm{u}-f)^2d\Omega},\\
&p(\bm{u};\bm{\alpha})\propto e^{-\lambda\int_\Omega\phi(\nabla \bm{u})d\Omega};
\end{split}
\end{equation*}
 thus, the first stage is to find a smooth approximation $\bm{u}$ by minimizing the multiphase generalizability (TS-MCMS) of (\ref{cai-eq}), which can be rewritten as
\begin{equation}\label{eq_feature_1a}
\min_{\bm{u} \in \text{W}^{1,2}(\Omega)}\left\{\int_{\Omega}(f-\mathcal{A} \bm{u})^{2} d\bm{x}+\mu\int_{\Omega}\phi(\nabla \bm{u}) d\bm{x}\right\},
\end{equation}
where $\mathcal{A}:\mathbb{R}^{d}\rightarrow \mathbb{R}$ is a convolutional blur operator, $\phi(\nabla \bm{u})=\nu|\nabla \bm{u}|^{2}+|\nabla \bm{u}|$ is a geometric prior of $\bm{u}$, and~$\mu=\frac{\lambda}{\gamma} $. Hence, this leads to the nonlinear system as
\begin{equation}\label{eq_feature_2}
\mathcal{F}(\bm{u};\bm{\alpha}):=\mathcal{A}^T\mathcal{A} \bm{u}-\mu\nabla\cdot(\phi^\prime(\nabla \bm{u}))=\bm{b},
\end{equation}
where $\bm{b}=\mathcal{A}^Tf$ and $\bm{\alpha}=(\mathcal{A}, \nabla, \mu,\nu)$.

\subsubsection{Feature Fusion~Segmentation} Once the features $\bm{u}$ are obtained, the~segmentation is performed by fusing $\bm{u}$ properly in the second stage; for~example, many novel image segmentation methods~\cite{XHCai2013,liu_WeightedVariational_2018,ma_ImageSegmentation_2018} have been proposed based on thresholding the smooth solution $\bm{u}$. Then, the fusing process for feature $\bm{u}$ is finished in (\ref{eq0a}).

The model-driven methods introduce prior knowledge
regarding many desirable mathematical
properties of the underlying anatomical
structure, such as phase field theory, $\Gamma$-approximation,
smoothness, and sparseness. The~informed priors may help to render
the segmentation method more robust and stable. However, these model-inspired methods generally solve
the optimization problem in the image domain, while the numerical minimization method for the feature representations $\bm{u}$ is very slow because the regularization of the TV-norm,~the high dimensionality of $\bm{u}$, as well as the nonlinear relationship between the images and the parameters pose significant computational challenges. Furthermore, it is challenging to introduce priors flexibly under different clinical scenes. These limitations make it hard for purely model-based segmentation to obtain the solutions efficiently and~flexibly.

The goal of this work was to learn powerful solvers of (\ref{eq_feature_2}) and (\ref{eq0a}) to aggregate a variety of mechanisms to
address the medical image segmentation problem~efficiently.

\subsection{Proposed Learnable Framework of TS-MCMS~Algorithm}
We summarized the two-stage algorithm to formulate medical image segmentation based on the TS-MCMS model, inspired by the CNN architectures of unrolled iterations, and we propose a learnable framework with two CNN modules on multiscale feature spaces, FAS-UNet (see Figure~\ref{Image_segmentation}b), aimed at learning the nonlinear inverse operators of (\ref{eq_feature_2}) and (\ref{eq0a}) in the context of the variational inverse problem to segment a given image~$f$.

It is already well known that the unrolled iterations of many classical algorithms can be considered as the layers of a neural network~\cite{larsson2017fractalnet,zhang2017polynet,gomez2017reversible,chen2018neural,Yang0LX16}. In~this part, we are not interested in designing another approach for inferring the classes in MgNet~\cite{He2019MgNetAU}, but~rather, we aim at extracting the features of a given image $f$.

Inspired by the variational segmentation model (\ref{eq_feature_1a}), one of the key ideas in the proposed architecture is that we split our framework into a solution module $\mathcal{T}_{\mathcal{K}}(f;\bm{\theta}_1)$ and a feature fusion module $\mathcal{S}_{\mathcal{K}}(\bm{u};\bm{\theta}_2)$, where $\mathcal{T}_{\mathcal{K}}(f;\bm{\theta}_1)$ is the feature extraction part of the framework (in the multi-stage case) and $\mathcal{S}_{\mathcal{K}}(\cdot;\bm{\theta}_2)$ is the stage fusion part to be learned.
Therefore, how to design the effective function maps $\mathcal{T}_{\mathcal{K}}$ for approximately solving (\ref{eq_feature_2}) and $\mathcal{S}_{\mathcal{K}}$ for approximating (\ref{eq0a}) is an important~problem.

This work applies a nonlinear multigrid method to design FAS-UNet for explainable medical image segmentation by learning the two following modules:
\begin{equation}\label{seg}
\left\{
\begin{aligned}
\bm{u} &= \mathcal{T}_{\mathcal{K}}(f;\bm{\theta}_1)\\
\bm{s} &= \mathcal{S}_{\mathcal{K}}(\bm{u};\bm{\theta}_2),
\end{aligned}
\right.
\end{equation}
where $f$ is an input image, $\bm{u}$ is the feature maps, and $\bm{s}$ is the prediction for the truth partitions, leading to the overall approximation function as
\begin{equation}\label{eq_seg}
\bm{s}= \mathcal{S}_{\mathcal{K}}(\mathcal{T}_{\mathcal{K}}(f;\bm{\theta}_1);\bm{\theta}_2),
\end{equation}
where $\bm{\theta}_1$ and $\bm{\theta}_2$ are parameters to be learned in the proposed explainable FAS-UNet\linebreak architecture.

To understand the approximation ability of the proposed modules $\mathcal{T}_{\mathcal{K}}(f;\bm{\theta}_1)$ and $\mathcal{S}_{\mathcal{K}}(\bm{u};\bm{\theta}_2)$ generated by the FAS-UNet architecture, we refer the readers to D. Zhou's work~\cite{ZHOU2020787}, which answers an open question in CNN learning theory about how deep CNN can be used to approximate any continuous function to an arbitrary accuracy when the depth of the neural network is large~enough.

\subsection{FAS-Module for Feature~Extraction}
In this part, we discuss how the multigrid method
can be used to solve nonlinear problems.
The Helmholtz Hypothesis~\cite{GHatfield2002} demonstrates that the extracted features can also be represented by solving the equation:
\begin{equation}
 \mathcal{F}(\bm{u},\bm{\alpha}) = \bm{b}:=\mathcal{A}^Tf,
 \label{linear_system}
\end{equation}
subject to
\begin{align*}\label{sub}
 \min_{\bm{u}} \|\mathcal{S}_{\mathcal{K}}(\bm{u};\bm{\theta}_2) - \bm{y} \|,
\end{align*}
where $\mathcal{F}$ denotes the transformation of combining feature $\bm{u}$ with a deblurred image $\bm{b}=\mathcal{A}^Tf$,
$\bm{u}$ is the unknown features, and $\bm{y}$ is the ground-truth of image $f$.
 Our starting point is the traditional FAS algorithm solving (\ref{linear_system}).

\subsubsection{The Full Approximation Scheme}\label{section_fas} The multigrid method is usually used to solve nonlinear algebraic systems (\ref{linear_system}). For~simplicity, the~parameter $\bm{\alpha}$ in $\mathcal{F}(\bm{u},\bm{\alpha})$ is omitted when only the classical FAS algorithm is involved, i.e.,
\begin{equation}
\mathcal{F}(\bm{u})= \bm{b}.
 \label{linear_system2}
\end{equation}

The multigrid ingredients including the error smoothing and the coarse grid correction ideas are not
restricted to the linear situation, but can be immediately used for the nonlinear problem itself, which leads to the so-called FAS algorithm. The~fundamental idea of the nonlinear multigrid is the same as in the linear case, and the~FAS method can be recursively defined
on the basis of a two-grid method. We start with the description of one fine--coarse cycle (finer grid layer $\ell$ and coarser grid layer $\ell+1$)
of the nonlinear two-grid method for solving (\ref{linear_system2}). To~proceed, let the fine grid equation be written as
\[\mathcal{F}^{\ell}(\bm{u}^{\ell}) = \bm{b}^{\ell}.\]

Firstly, we compute an approximation $\bar{\bm{u}}^{\ell}:=\bm{u}_{m}^{\ell}$ of the fine grid problem by applying $m$ pre-smoothing steps to $\bm{u}^{\ell}$ as follows
\begin{algorithmic}
\STATE $\bm{u}_{0}^{\ell}=\bm{u}^{\ell}$;
 \FOR{$k=1: m$}
 \STATE
$\bm{u}_k^{\ell}=\bm{u}_{k-1}^{\ell}+(\mathcal{F}^{\ell})^\prime(\bm{b}^{\ell}-\mathcal{F}^{\ell}(\bm{u}_{k-1}^{\ell})),$
 \ENDFOR
 \STATE $\bar{\bm{u}}^{\ell}=\bm{u}_m^{\ell}$,
\end{algorithmic}
which can be obtained via solving the least-squares problem defined by
\[
\min\limits_{\bm{u}}\left\{\mathcal{E}(\bm{u}):=\frac{1}{2}\|\mathcal{F}^{\ell}(\bm{u})-\bm{b}^{\ell}\|^2\right\}.
\]

Secondly, the~errors to the solution have to be smoothed such that they can be approximated on a coarser grid. Then, the defect $\bm{r}^{\ell}=\bm{b}^{\ell}-\mathcal{F}^{\ell}(\bar{\bm{u}}^{\ell})$ is computed, and~an analog of the linear defect equation is transferred to the coarse grid, which is defined by
\begin{equation}\label{coarse_eq}
\mathcal{F}^{\ell+1}(\bm{u}^{\ell+1}) = \bm{b}^{\ell+1}:=I_{\ell}^{\ell+1}\bm{r}^{\ell}+\mathcal{F}^{\ell+1}(I_{\ell}^{\ell+1} \bar{\bm{u}}^{\ell}).
\end{equation}

The coarse grid
corrections are interpolated back to the fine grid by
\begin{equation}\label{coarse_correction}
\bar{\bm{u}}^{\ell} \leftarrow \bar{\bm{u}}^{\ell}+I_{\ell+1}^{\ell}\left[\bm{u}^{\ell+1}-I_{\ell}^{\ell+1} \bar{\bm{u}}^{\ell}\right],
\end{equation}
where $\bm{u}^{\ell+1}$ is a solution of the coarse grid equations, then the errors are finally post-smoothed~by
\begin{algorithmic}
\STATE $\bm{u}_{0}^{\ell}=\bar{\bm{u}}^{\ell}$;
 \FOR{$j=1: m$}
 \STATE
$\bm{u}_j^{\ell}=\bm{u}_{j-1}^{\ell}+(\mathcal{F}^{\ell})^\prime(\bm{b}^{\ell}-\mathcal{F}^{\ell}(\bm{u}_{j-1}^{\ell})),$
 \ENDFOR
\STATE $\hat{\bm{u}}^{\ell}=\bm{u}_m^{\ell}$.
\end{algorithmic}

This means that, once the solution of the fine grid problem is obtained, the~coarse grid correction does not introduce any changes via interpolation. We regard this property as an essential one, and~in our derivation of the coarse grid optimization problem, we make sure that it is~satisfied. 

\begin{figure*}[h]
  \centering
  \includegraphics[width=0.98\linewidth]{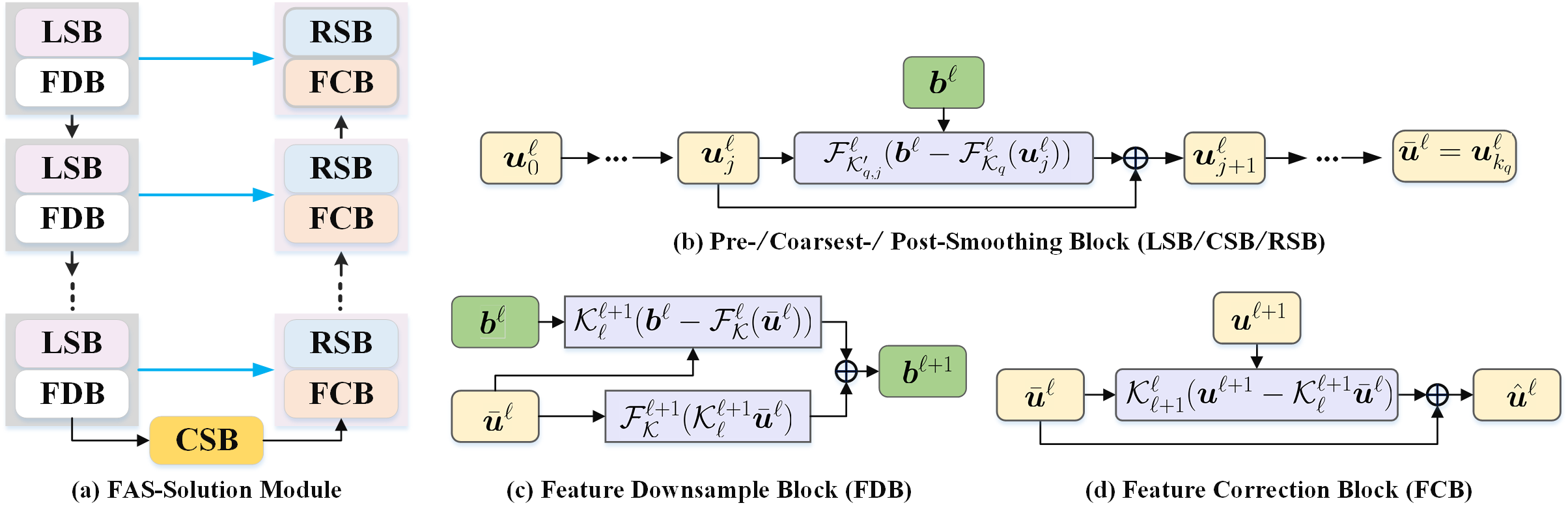}
    \caption{The overall flowchart of the proposed feature extraction module (FAS-Solution module) with multi-grid architecture. It consists of three major ingredients, i.e. pre-/coarsest-/post-smoothing blocks (LSB/CSB/RSB), feature downsample block (FDB), and feature correction block (FCB).}
\label{FAS_net}
\end{figure*}

\subsubsection{FAS-Solution Module---A Learnable Architecture for FAS~Solution} In this part, we unroll the multiscale correction process of the multigrid method and design a series of deep FAS-Solution modules to
propagate the features of input image $f$. Figure~\ref{FAS_net} demonstrates the cascade of all ingredients at
each FAS iteration of our propagative~network.

We now consider a decomposition of the image representation into partial sums,
which correspond to the multiscale feature sequence, having the following idea in mind. To~learn features that are invariant to noise and uninformative
intensity variations, we propose a generative feature
module $\bm{u}= \mathcal{T}_{\mathcal{K}}(f;\bm{\theta}_1)$ allowing for a significant reduction of the number of parameters involved, which involves a learnable FAS update for solving the nonlinear system (\ref{linear_system}). Next, we analyze its three key components as follows.

\textbf{Pre-/coarsest-/post-smoothing block:} Error
smoothing is one of the two basic principles of the FAS approach. At~each pre-smoothing (or coarsest-processing or post-processing) step, to~establish an efficient error correction and reduce
the computational costs, we propose
to generate the error-based iterative scheme. The~main motivation of the learnable pre-/coarsest-/post-smoothing blocks (LSB/CSB/RSB) is to provide another way to robustly resolve the ambiguity
of the feasible solutions. Therefore, we further unroll the Newton update process for calculating the approximate solution of the feature maps, then design a series of deep pre-/coarsest-/post-smoothing blocks as
\begin{algorithmic}
 \FOR{$j=0: k_q-1$}
 \STATE
$\begin{aligned}
\bm{u}_{j+1}^{\ell}=&\bm{u}_j^{\ell}+\mathcal{M}(\bm{u}_j^{\ell};\mathcal{K}_{q,\ell},\mathcal{K}^\prime_{q,\ell,j},\bm{b}^{\ell})\\
=&\bm{u}_j^{\ell}+\mathcal{F}_{\mathcal{K}^\prime_{q,j}}^\ell(\bm{b}^{\ell}-\mathcal{F}^\ell_{\mathcal{K}_{q}}(\bm{u}_j^{\ell})),
\end{aligned}$
 \ENDFOR
 \STATE $\bar{\bm{u}}^{\ell}=\bm{u}_{k_q}^{\ell}$;
\end{algorithmic}
where the residual network block $\mathcal{M}(\bm{u}^{\ell};\mathcal{K},\mathcal{K}^\prime,\bm{b}^{\ell})$ is a trainable feature correction network; here, $\ell=1,\dots, \mathcal{L}$ with $\mathcal{L}$-grid cycles. The~above deep smoothing series denotes the pre-smoothing block when $q:=l$, the~$\mathcal{L}^{th}$ coarsest-smoothing when $q:=m$, and the post-smoothing when $q:=r$. $\mathcal{F}^\ell_{\mathcal{K}_{q,j}^\prime}$ (or $\mathcal{F}_{\mathcal{K}_{q}}^\ell$, which means that the convolution $\mathcal{K}_{q,\ell}$ will share the same weights in the overall pre-/post-precessing smoothing steps of the $\ell^{th}$ grid cycle) represents operations consisting of three main components, including convolution $\mathcal{K}_{q,\ell,j}^\prime$ (or $\mathcal{K}_{q,\ell}$) with $p$ filters, ReLU function $\varphi(\cdot)$, and batch normalization $\psi(\cdot)$, such that $\mathcal{F}_{\mathcal{K}^\prime_{q,j}}^\ell(\cdot):=\psi(\varphi(\mathcal{K}^\prime_{q,\ell,j}(\cdot)))$. Especially, $\bm{b}^{0}:=\mathcal{K}^0f$ is an initial feature in the finest grid, which is obtained by learning the convolution $\mathcal{K}^0$ with $p$ filters.

\textbf{Feature downsample block:}
The choice of restriction and interpolation operators $I_{\ell}^{\ell+1}$ and $I_{\ell+1}^{\ell}$ in the FAS algorithm, for~the intergrid transfer of grid functions, is closely related to the choice of the coarse strategy. Here, we design the learnable convolution for transfer
operators, i.e.,~the grid transfers between the finer grid $\ell$
and the coarser grid $\ell+1$.

The low-frequency components represent meaningful image features on a coarse grid $\ell+1$, whereas the high-frequency components do not because they are not ``visible'' on the coarse grid, which means that the frequency information on the coarse grid can be extracted from the right-side term defined by
\begin{equation}\label{coar}
\bm{b}^{\ell+1}=\mathcal{K}_{\ell}^{\ell+1}(\bm{b}^{\ell}-\mathcal{F}^{\ell}_\mathcal{K}(\bar{\bm{u}}^{\ell}))+\mathcal{F}^{\ell+1}_\mathcal{K}(\mathcal{K}_{\ell}^{\ell+1} \bar{\bm{u}}^{\ell}),
\end{equation}
where $\bm{b}^{\ell}$ and $\bar{\bm{u}}^{\ell}$ are the inputs of the downsample block and $\bm{b}^{\ell+1}$ is the output of the downsampling module in the feature space; here, $\ell=1,\dots, \mathcal{L}-1$ with $\mathcal{L}$-grid cycles. Similar to $\mathcal{F}_{\mathcal{K}_j}^\ell$, $\mathcal{K}_{\ell}^{\ell+1}$ is a learnable downsample operation that would be used to approximate the restriction function $I_{\ell}^{\ell+1}$ in (\ref{coarse_eq}) or (\ref{coarse_correction}), such as convolution with a stride of $2$ and $p$ filters. $\mathcal{F}^{\ell}_\mathcal{K}$ and $\mathcal{F}^{\ell+1}_\mathcal{K}$ are the nonlinear convolutional blocks in the fine and coarse layers, respectively. In~general, $\mathcal{F}_{\mathcal{K}}^\ell$ denotes the operator consisting of three main components, including convolution $\mathcal{K}_{\ell}$ with $p$ filters, ReLU function $\varphi(\cdot)$, and batch normalization $\psi(\cdot)$, such that $\mathcal{F}_{\mathcal{K}}^\ell(\cdot):=\psi(\varphi(\mathcal{K}_{\ell}(\cdot)))$. Note that $\bm{b}^{\ell}-\mathcal{F}^{\ell}_\mathcal{K}(\bar{\bm{u}}^{\ell})$ is equivalent to the residual of the images in the fine layer, then $\mathcal{F}^{\ell+1}_\mathcal{K}(\mathcal{K}_{\ell}^{\ell+1} \bar{\bm{u}}^{\ell})$ is added to reduce the loss of image information compared with directly pooling the image. The~feature downsample block (FDB) architecture is shown in Figure~\ref{FAS_net}b.

\textbf{Feature correction block:} The purpose of the feature correction block (FCB) is to take the detailed information extracted from the coarser grid into account and help to compensate the encoded features $\bar{\bm{u}}^{\ell}$.
 The coarse grid
corrections are interpolated back to the fine grid,~i.e.,
\begin{equation}\label{coarse_co}
\hat{\bm{u}}^{\ell} \leftarrow \bar{\bm{u}}^{\ell}+\mathcal{K}_{\ell+1}^{\ell}\left[\bm{u}^{\ell+1}-\mathcal{K}_{\ell}^{\ell+1} \bar{\bm{u}}^{\ell}\right],
\end{equation}
where $\mathcal{K}_{\ell+1}^{\ell}$ is a learnable upsampling operation that would be used to approximate the interpolation function $I_{\ell+1}^{\ell}$ in (\ref{coarse_correction}), such as the transposed convolution with a stride of $2$ and $p$ filters; here, $\ell=1,\dots, \mathcal{L}-1$ with $\mathcal{L}$-grid cycles. Obviously,~$\bm{e}^{\ell+1}=\bm{u}^{\ell+1}-\mathcal{K}_{\ell}^{\ell+1} \bar{\bm{u}}^{\ell}$ is the residual features on the coarse grid. Compared to directly upsampling $\bm{u}^{\ell+1}$, the~transposed convolution $\mathcal{K}_{\ell+1}^{\ell}\bm{e}^{\ell+1}$ of the residual feature maps $\bm{e}^{\ell+1}$ is used as the error corrections to update the fine grid approximation $\hat{\bm{u}}^{\ell}$, which will compensate the information of feature maps $\bar{\bm{u}}^{\ell}$. Such a transposed convolution could learn a self-adaptive mapping to restore features with more detailed~information.

Based on these designs for nonlinear operator $\mathcal{F}_{\mathcal{K}}^\ell$ and~two grid
transfer convolutions $\mathcal{K}_{\ell}^{\ell+1}$ and $\mathcal{K}_{\ell+1}^{\ell}$ with $p$ filters, we aimed to approximate the feature solution of (\ref{eq_feature_1a}) by learning these feature extraction parameters as
\vspace{-9pt}
\begin{equation}
\begin{split}
\bm{\theta}_1=\Big\{\big(\mathcal{K}_{\ell+1}^{\ell},&\mathcal{K}_{\ell}^{\ell+1},(\mathcal{K}_{q,\ell}),(\mathcal{K}_{q,\ell,j}^\prime)_{j=1}^{k_q},\mathcal{K}_\ell\big)_{\ell=1}^{\mathcal{L}-1},\\ &\mathcal{K}^0, (\mathcal{K}_{m,{\mathcal{L}}}),(\mathcal{K}_{m,{\mathcal{L}},j}^\prime)_{j=1}^{k_m}\Big|q\in\{l,r\}\Big\}
\end{split}
\label{model_parameter_1}
\end{equation}
in $\bm{u} = \mathcal{T}_{\mathcal{K}}(f;\bm{\theta}_1)$, thus further improving image~segmentation.

\subsection{Learning Feature Fusion~Segmentation}
It is well known that, in~the segmentation task, each pixel is labeled as either $0$ or $1$ so that organ pixels can be accurately identified within the tight bounding box.
In the second stage of the two-stage multiphase variational image segmentation (\ref{eq_feature_1a}), the~traditional method is that users manually set one or more thresholds according to their professional prior, with~all pixels in the same object sharing the threshold, and~then, filter the feature to obtain the segmentation result. Another method is to obtain the final segmentation result by $k$-means clustering (the number of categories is given artificially, and the~initial clustering center is adjusted continuously during the clustering process) \cite{XHCai2013}. This approach leads to a large amount of computation (recalculation of the metrics for each iteration) and unstable segmentation results (only considering the relationship between pixels and centers, not the relationship between pixels).

The second key component of our proposed FAS-UNet framework is how to design the segmentation module $\mathcal{S}_{\mathcal{K}}(\bm{u};\bm{\theta}_2)$ to compute segmented mask $\bm{s}$. However, the~fusion segmentation module takes a batch of multiscaled features from the FAS module as the input and~outputs the mass segmentation masks. Finally, the~pixel segmentation computes the mapping from smaller-scale possibility predictions to binary~masks.

Based on this idea, the~feature fusion segmentation module is constructed, which comprises a convolutional operation and an activation function. Intuitively, the~module $\mathcal{T}_{\mathcal{K}}(f;\bm{\theta}_1)$ of the feature extraction based on deep learning obtains the multi-channel feature maps $\bm{u}$ (much larger than the number of categories) in the first stage. Then, a~shallow convolutional network is constructed to learn the parameters $\mathcal{K}_{p}$ corresponding to the mapping $\rho(\mathcal{K}_{p}(\cdot)): \mathbb{R}^p\rightarrow \mathbb{R}^c$ from the feature maps $\bm{u}$ to the segmentation probability maps through \emph{softmax} function $\rho(\cdot)$, which improves the traditional practice and has better~generalization.

Based on these designs for channel transfer convolution $\mathcal{K}_{p}$ with $c$ filters ($c$ is the number of segmentation categories), we aimed to approximate the final multiphase segmentation probability maps $\mathcal{S}(\cdot;\bm{\beta})$ of (\ref{eq_feature_1a}) by learning these fusion parameters as
\begin{equation}
\bm{\theta}_2=\{\mathcal{K}_{p}\}
\label{model_parameter_2}
\end{equation}
in $\bm{s} = \mathcal{S}_{\mathcal{K}}(\bm{u};\bm{\theta}_2)$, thus further refining the segmentation~mask.

\subsection{Loss~Function}

The proposed FAS-UNet architecture {can be rewritten} as
\[\bm{s}= \mathcal{S}_{\mathcal{K}}(\mathcal{T}_{\mathcal{K}}(f;\bm{\theta}_1);\bm{\theta}_2),\]
{which} requires the loss function $\mathcal{L}(\bm{\theta};\mathcal{D}_{\text {train }})$ to optimize the model parameters $\bm{\theta}:=\{\bm{\theta}_1,\bm{\theta}_2\}$. It can measure the error between the prediction and labels, and~the gradients of the weights in the loss function can be back-propagated to the previous layers in order to update the model~weights.

To {proceed}, we considered the training data $\mathcal{D}_{\text {train }}=\left\{\left(f_i, y_i\right)\right\}^n_{i=1}$ from a set of classes $\mathcal{C}_{\text {train }}=\{0, \dots, c-1 \}^d$ used for training a pixel classifier, where $f_i$ is an image sample, $y_i \in \mathcal{C}_{\text {train }}$ is the corresponding label, $c$ is the number of object categories segmented in the datasets, $d$ indicates the number of image pixels, and $n$ denotes the number of training samples. We employed the cross-entropy as the loss function, leading to the optimization problem as follows:
\begin{equation}\label{ce}
\begin{split}
\min\limits_{\bm{\theta}}\Bigg\{\mathcal{L}(\bm{\theta};\mathcal{D}_{\text {train }}):= &\sum_{i=1,j=1}^{n,d}\Big(\bm{\bar{s}}_i^{(j)} \cdot \log(\bm{s}_i^{(j)} )\\& + (\bm{1}-\bm{\bar{s}}_i^{(j)})\cdot \log(\bm{1}-\bm{s}_i^{(j)}) \Big)\Bigg\},
\end{split}
\end{equation}
where $\bm{s}_i^{(j)}$ denotes the predicted probabilistic vector of the $j^{th}$ pixel in the $i^{th}$ sample and $\bm{\bar{s}}_i^{(j)}$ corresponds to the one-hot-encoded label of the ground-truth $y_i^{(j)}$ at the $j^{th}$ pixel in the $i^{th}$ sample. $\bm{s}_i^{(j)}\in \mathbb{R}^c$. Finally, the~predicted class of the $j^{th}$ pixel of the $i^{th}$ image would be given by
\[
y_i^{(j)} = \mathop{\arg\max}\limits_{k}\{\bm{s}_i^{(j)}(1), \dots,\bm{s}_i^{(j)}(k), \dots, \bm{s}_i^{(j)}(c)\},
\]
where $\bm{s}_i^{(j)}(k)\in \mathcal{C}_{\text {train }}$.

\section{Datasets and experiments}\label{sect3}
In this section, we first introduce the evaluation metrics of medical image segmentation, and also describe the datasets and the experimental settings that we use for 2D CT image segmentation and 3D medical volumetric segmentation. Next, we analyze the sensitivity of 2D FAS-Unet to each hyperparameter configuration by a series of experiments. Finally, we evaluate the effectiveness of the proposed 2D/3D FAS-Unet through comparative experiments.

\subsection{Evaluation metrics}
There are many metrics to quantitatively evaluate segmentation accuracy, each of which focuses on different aspects. In this work, we employ average Dice similarity coefficient (a-DSC), average precision (a-Preci), and average symmetric surface distance (a-SSD), which are widely used in the segmentation task as evaluation metrics to evaluate the performance of the model. The a-DSC/a-Preci/a-SSD are calculated by averaging the DSC/Preci/SSD of each category \cite{taha2015metrics}.

Dice score is the most used metric in validating medical image segmentation, also called DSC score \cite{Dice1945Measures} defined by
\begin{equation*}\label{dice}
  \textbf{DSC}(S,Y) = 2 \times \frac{|S\bigcap Y|}{|S| + |Y|},
\end{equation*}
where $S$ and $Y$ denote the automatically segmentation set of image and the manually annotated ground truth, respectively. $|\cdot|$ denotes the measure of a set.
The above formulas computes overlap between the prediction and the ground-truth, to evaluate overall effect of the segmented results. However, it is fairly insensitive to the precise boundary of the segmented regions. Precision effectively describes the purity of the prediction relative to the ground-truth, or measures that the number of those pixels actually has a matching ground truth annotation by calculating
\begin{equation*}
  \textbf{Preci} = \frac{TP}{TP + FP},
\end{equation*}
where a true positive (TP) is observed when a prediction-target mask pair has a score which exceeds some predefined threshold, and a false positive (FP) indicates a predicted object mask has no associated ground truth. The SSD value between two finite point sets $S$ and $Y$ is defined as follows
\begin{equation*}
  \textbf{SSD}(\partial S, \partial Y) = \frac{\sum\limits_{s \in \partial S} d(s, \partial Y) + \sum\limits_{y \in \partial Y}d(y, \partial S)}{|\partial S| + |\partial Y|},
\end{equation*}
where $d(v, X) = \min_{x \in X} \| v -x \| $  denotes the minimum Euclidean distance from point $v$ to all points of $X$.

\subsection{Datasets  and experimental setup}
We evaluate the proposed method and other state-of-the-art methods in 2D SegTHOR datasets 
\cite{segthor}, 3D HVSMR-2016 datasets 
\cite{pace2015interactive} and 3D CHAOS-CT datasets 
\cite{kavur2020chaos, CHAOSdata2019}, respectively. We introduce their details and data processing methods as follows.

\subsubsection{Data Preparation.}
For the SegTHOR datasets, in~order to reduce the GPU memory cost and reduce the image noise, we first split the original 3D data into many 2D images along the Z-axis. Secondly, we used $\bm{f} = \bm{I}[96:400, 172:396]$ as the input image, where $\bm{I}$ denotes the original 2D slice. We removed the slices of the pure black ground-truth when it was in training on the 
datasets. 

For the 3D HVSMR-2016 datasets, we directly used the sliding window cropping method with strides of $64 \times 64 \times 32$ to crop the volumes. In~general, before~cropping the 3D whole-volume into several overlapping sub-volumes of size $128 \times 128 \times 64$, a~$(32,32,16)$-voxel padding with zero filling was first added to each direction of the 3D whole-volume. Then, after~these
operations, all remaining sub-volumes whose sizes were smaller than $128 \times 128 \times 64$ were resized to $128 \times 128 \times 64$ with zero-filling,
and the intensity values of all patches were in $[0, 4808]$.

For the 3D CHAOS-CT datasets, which were used as for the liver segmentation experiments, we first cropped the volumes in the $x, y$ directions to obtain an ROI with a size of $380 \times 440 \times z$ and~then used the above sliding window cropping method to crop out several 3D sub-volumes,
where those intensity values were in $[-1200, 1096]$.

Although the noise problem can be improved by data pre-processing, our aim was not to pursue the best performance of the network on these datasets; we compared the performance of each method under fairer conditions. Using some data pre-processing techniques may be particularly beneficial to some methods, while at the same time, they may degrade the performance of others, so we did not use more complex data pre-processing~techniques.

\subsubsection{Experimental configurations.} We used mini-batch stochastic gradient descent (SGD)
 to optimize the proposed model, in~which the initial learning rate, momentum parameter, and~weight decay parameter were set to 0.01, 0.99, and~$10^{-4}$, respectively. We set the batch size as 16, 4, and~4 for the 2D SegTHOR datasets, 3D HVSMR-2016 datasets, and~3D CHAOS-CT datasets, respectively. The~maximum epochs of the three datasets were set to 150, 150, and~300, respectively. We used also the decay strategy to update the learning rate.
 The network initialization method was defined as Kaiming initialization, and~the activation function was set as ReLU. The~numbers of grid cycles for 2D and 3D FAS-UNet were $\mathcal{L}=5$ and $\mathcal{L}=4$, respectively. The~kernel sizes of the 2D and 3D networks were set to $3 \times 3$ and $3 \times 3\times 3$ as the defaults, respectively. We did not use the weight-sharing scheme for the convolution $\mathcal{K}_{q,\ell,j}^\prime$ (within the outer-level nonlinear operator $\mathcal{F}_{\mathcal{K}^\prime_{q,j}}^\ell$) within one smoothing block.
 Table~\ref{fasunet} shows the details of the 2D FAS-UNet framework, and~3D FAS-UNet has a similar architecture, except for replacing the 2D convolution with a 3D convolution.

\begin{table*}
\begin{center}
\caption{A standard configuration of the proposed 2D FAS-Unet.}
    \setlength{\tabcolsep}{0.6mm}{
\begin{tabular*}{\hsize}{@{\extracolsep{\fill}}ccc|ccc}
\toprule[1.5pt]
\multicolumn{1}{c|}{ \textbf{Name}} & \multicolumn{1}{c|}{\textbf{2D operation}} & \multicolumn{1}{c|}{\textbf{Output size}} 
& \multicolumn{1}{c|}{ \textbf{Name}} & \multicolumn{1}{c|}{\textbf{2D operation }}& \multicolumn{1}{c}{\textbf{Output size}} \\ 
\midrule[0.8pt]
\textbf{Input} & {$f$,\quad initialization $\bm{u}_0^1$} &
$p \times H \times W $ & {}& {}                                                                                            &   \\ 
\hline
\textbf{LSB1}         & {\begin{tabular}[c]{@{}c@{}} $
\begin{pmatrix}  \text{conv3, s=1} \\      \text{ReLU+BN}\\ \text{ conv3, s=1} \\ \text{ ReLU+BN}
\end{pmatrix} \times k_l $
\end{tabular}} & $p \times H \times W $    & \textbf{RSB1}   & {\begin{tabular}[c]{@{}c@{}}$
\begin{pmatrix}   \text{conv3, s=1} \\      \text{ReLU+BN}\\ \text{ conv3, s=1} \\ \text{ ReLU+BN}
\end{pmatrix} \times k_r $\end{tabular}}  & $p \times H \times W $      \\ 
\hline
\textbf{FDB1}  & {\begin{tabular}[c]{@{}c@{}}conv3, s=2 for $\bm{f}$\\      conv3, s=2 for $\bm{u}$\end{tabular}} & $p \times \dfrac{H}{2} \times \dfrac{W}{2} $  &
\textbf{FCB1}   & {deconv3, s=2} & $p \times H \times W $      \\ 
\hline
\textbf{LSB2}         &{\begin{tabular}[c]{@{}c@{}}$
\begin{pmatrix}   \text{conv3, s=1} \\      \text{ReLU+BN}\\ \text{ conv3, s=1} \\ \text{ ReLU+BN}
\end{pmatrix} \times k_l $ \end{tabular}} & $p \times \dfrac{H}{2} \times \dfrac{W}{2} $   & \textbf{RSB2}   & {\begin{tabular}[c]{@{}c@{}}$
\begin{pmatrix}   \text{conv3, s=1} \\      \text{ReLU+BN}\\ \text{ conv3, s=1} \\ \text{ ReLU+BN}
\end{pmatrix} \times k_r $\end{tabular}} & $p \times \dfrac{H}{2} \times \dfrac{W}{2}$   \\ 
\hline
\textbf{FDB2}  &  {\begin{tabular}[c]{@{}c@{}}conv3, s=2 for $\bm{f}$\\      conv3, s=2 for $\bm{u}$\end{tabular}}           & $ p \times \dfrac{H}{4} \times \dfrac{W}{4} $   &
\textbf{FCB2}   & {deconv3, s=2}                                                                              &$ p \times \dfrac{H}{2} \times \dfrac{W}{2} $ \\ 
\hline
\textbf{LSB3}         & {\begin{tabular}[c]{@{}c@{}}$
\begin{pmatrix}  \text{conv3, s=1} \\      \text{ReLU+BN}\\     \text{ conv3, s=1} \\ \text{ ReLU+BN}
\end{pmatrix} \times k_l $ \end{tabular}} & $p \times \dfrac{H}{4} \times \dfrac{W}{4} $   & \textbf{RSB3}   &{\begin{tabular}[c]{@{}c@{}}$
\begin{pmatrix}  \text{conv3, s=1} \\      \text{ReLU+BN}\\ \text{ conv3, s=1} \\ \text{ ReLU+BN}
\end{pmatrix} \times k_r $\end{tabular}}& $p \times \dfrac{H}{4} \times \dfrac{W}{4} $   \\ 
\hline
\textbf{FDB3}         & {\begin{tabular}[c]{@{}c@{}}conv3, s=2 for $\bm{f}$\\      conv3, s=2 for $\bm{u}$\end{tabular}}             & $p \times \dfrac{H}{8} \times \dfrac{W}{8} $   & \textbf{FCB3}   & {deconv3, s=2}                                                                              & $p \times \dfrac{H}{4} \times \dfrac{W}{4} $  \\ 
\hline
\textbf{LSB4}         & {\begin{tabular}[c]{@{}c@{}}$
\begin{pmatrix}  \text{conv3, s=1} \\      \text{ReLU+BN}\\ \text{ conv3, s=1} \\ \text{ ReLU+BN}
\end{pmatrix} \times k_l $\end{tabular}} & $p \times \dfrac{H}{8} \times \dfrac{W}{8} $   & \textbf{RSB4}   & \multicolumn{1}{c}{\begin{tabular}[c]{@{}c@{}}$
\begin{pmatrix}  \text{conv3, s=1} \\      \text{ReLU+BN}\\ \text{ conv3, s=1} \\ \text{ ReLU+BN}
\end{pmatrix} \times k_r $\end{tabular}} & $p \times \dfrac{H}{8} \times \dfrac{W}{8} $   \\ 
\hline
\textbf{FDB4}         & {\begin{tabular}[c]{@{}c@{}}conv3, s=2 for $\bm{f}$\\      conv3, s=2 for $\bm{u}$\end{tabular}}                                    & $p \times \dfrac{H}{16} \times \dfrac{W}{16} $   & \textbf{FCB4}   & {deconv3,   s=2}                                                                              & $p \times \dfrac{H}{8} \times \dfrac{W}{8} $  \\ 
\hline
\textbf{CSB}         & \multicolumn{4}{c}{
\begin{tabular}[c]{@{}c@{}}$
\begin{pmatrix}   \text{conv3, s=1} \\      \text{ReLU+BN}\\ \text{ conv3, s=1} \\ \text{ ReLU+BN}
\end{pmatrix} \times k_m $ \end{tabular}
} & $ p \times \dfrac{H}{16} \times \dfrac{W}{16} $   \\ 
\bottomrule[1.5pt]
\end{tabular*}
}
\label{fasunet}
\end{center}
\end{table*}

\subsubsection{Parameter complexity.} To compute the number of parameters of the proposed model, we first denote the number of parameters of 2D convolution kernel $\mathcal{K}_{_{2d}}$ with the shape $p \times p \times k_c \times k_c $ as follows:
\[\eta(\mathcal{K}_{_{2d}}) = p^2 N=p^2(k_c)^2. \]
Similarly, the~number of parameters of 3D convolution $\mathcal{K}_{_{3d}}$ with the shape $rp \times rp \times k_c \times k_c\times k_c $ is defined as
\[\eta(\mathcal{K}_{_{3d}}) = r^2 p^2 (k_c)^3. \] 

 Especially if the channel ratio satisfies that $r<\frac{\sqrt{3}}{k_c}$, one has $\eta(\mathcal{K}_{_{2d}})>\eta(\mathcal{K}_{_{3d}})$. Thus, the~number of parameters of convolution kernel set $\theta_1$ in the proposed FAS-UNet can be computed by 
\[\begin{aligned}
 \eta(\theta_1)
 & =cp (k_c)^2 + {\Big(} (k_m + 1)+ (\mathcal{L}-1)(k_l + k_r+5) \Big)\eta(\mathcal{K}) \\
 & \approx \left((k_m + 1) + (\mathcal{L}-1)(k_l + k_r+5)\right)\eta(\mathcal{K}),
\end{aligned}
\]
where $c$ is the channel number of the input image and $\eta(\mathcal{K})$ is the number of parameters of each 2D or 3D convolution $\mathcal{K}$.

If setting $k_c=3$, $p=64$, $\{k_l, k_m, k_r\}= \{3,7, 4\}$ and $\mathcal{L}=5$, one has $\eta(\mathcal{K}_{_{2d}})=36864$; hence, 2D FAS-UNet has approximately $\eta(\theta_1)=56\eta(\mathcal{K}_{_{2d}})=2064384$ parameters. In~addition, one also has $\eta(\mathcal{K}_{_{3d}}) = 27648 $ when $k_c=3, p=32$. If~setting $\{k_l, k_m, k_r\}=\{3, 5, 2\}$ and $\mathcal{L}=4$, thus 3D FAS-UNet has approximately $\eta(\theta_1)=36\eta(\mathcal{K}_{_{3d}})=995328$ parameters. Here, we did not compute $\eta(\theta_2)$, where a small amount of parameters are~involved.

Our experiments were implemented on the PyTorch framework and two NVIDIA Geforce RTX 2080Ti GPUs with 11GB memory.
For each quantitative result in the experiments, we repeated the experiment twice and chose the best one to compute the mean/std. Note that we used the same pipeline for all these experiments of each dataset for a fair comparison. The~networks under comparison were trained from~scratch.

\subsection{Ablation~Studies}
We conducted four groups of ablation studies on the 2D SegTHOR datasets to optimize the hyperparameter configurations of the proposed~framework.

\begin{table*}
\begin{center}
\caption{Quantitative assessment with a-DSC, a-SSD and a-Preci values of different pre-/coarest-/post- smoothing iteration numbers $\{k_l,k_m,k_r\}$ using the proposed 2D FAS-Unet framework on the 2D SegTHOR validation datasets with four organs: esophagus, heart, trachea and aorta. "Mean" denotes an average score segmenting all organs. The best and second places are highlighted in Bold font and underlined ones, respectively.}
    \setlength{\tabcolsep}{0.8mm}{
\begin{tabular*}{\hsize}{@{\extracolsep{\fill}}clcccccc}
\toprule[1.5pt]
\multicolumn{2}{c}{\textbf{Number of smoothing steps} } & \multicolumn{3}{c}{\textbf{Fixing} $k_m$ \textbf{and} $k_r$\textbf{, varying} $k_l$ }&\multicolumn{3}{c}{\textbf{Fixing} $k_l$ \textbf{and} $k_m$\textbf{, varying} $k_r$ }    \\
\cmidrule[0.7pt](lr){3-5} \cmidrule[0.7pt](lr){6-8}
\multicolumn{2}{c}{ } & \textbf{{\{}2,7,2{\}}}    & \textbf{{\{}3,7,2{\}}}    & \textbf{{\{}4,7,2{\}}}    & \textbf{{\{}3,7,3{\}}}    & \textbf{{\{}3,7,4{\}}}   & \textbf{{\{}3,7,5{\}}}    \\
\midrule[0.8pt]
\multicolumn{2}{c}{\textbf{Params}}        & \textbf{0.41M}   & \underline{0.44M}  & 0.48M    & \textbf{0.48M }  & \underline{ 0.52M }            & 0.56M              \\
\midrule[0.8pt]
  & \textbf{Esophagus}      & \underline{ 73.56 $\pm$ 9.58} & \textbf{74.55 $\pm$ 8.83}                & 71.82 $\pm$ 9.42                        & \underline{ 74.06 $\pm$   9.80} & \textbf{74.24 $\pm$ 8.01}               & 73.19 $\pm$ 9.79                        \\
& \textbf{Heart}          & \textbf{93.63 $\pm$ 2.22}                                       & 92.96 $\pm$ 3.46                         & \underline{ 93.21 $\pm$ 2.28} & \underline{ 92.98 $\pm$ 1.55}   & \textbf{94.22 $\pm$ 1.52}               & 92.64 $\pm$ 4.72               \\
& \textbf{Trachea}        & 83.06 $\pm$ 4.33                                                & \textbf{85.78 $\pm$ 4.32}                & \underline{ 84.22 $\pm$ 4.04} & \textbf{86.43 $\pm$ 4.01}                 & 84.47 $\pm$ 5.33                        & \underline{ 84.54 $\pm$ 6.81} \\
& \textbf{Aorta}          & \underline{ 89.78 $\pm$ 4.66} & 89.09 $\pm$ 4.9                          & \textbf{89.96 $\pm$ 5.78}               & 88.27 $\pm$ 5.13                          & \underline{ 89.39 $\pm$ 7.26} & \textbf{92.11 $\pm$ 1.36}               \\
\multirow{-5}{*}{\textbf{a-DSC(\%)}}        
& \textbf{Mean}           & \underline{ 85.01}            & \textbf{85.60}                           & 84.80                                   & 85.44                                     & \underline{ 85.58}            & \textbf{85.62}                          \\
\midrule[0.8pt]
 & \textbf{Esophagus}      & 4.16 $\pm$ 1.91                                                 & \textbf{2.67 $\pm$ 0.98}                 & \underline{ 3.35 $\pm$ 1.34}  & 3.22 $\pm$ 1.14                           & \textbf{2.67 $\pm$ 0.79}                & \underline{ 2.69 $\pm$ 1.06}  \\
& \textbf{Heart}          & \textbf{4.61 $\pm$ 3.97}                                        & \underline{ 12.48 $\pm$ 16.53} & 16.54 $\pm$ 24.38                       & \underline{ 2.66 $\pm$ 0.94}    & \textbf{1.94 $\pm$ 0.66}                & 18.5 $\pm$ 27.98                        \\
& \textbf{Trachea}        & \underline{ 5.22 $\pm$ 2.37}                          & \textbf{3.18 $\pm$ 1.15}                 & 6.61 $\pm$ 4.57                         & \underline{ 4.80 $\pm$ 4.70}    & \textbf{3.01 $\pm$ 2.48}                & 6.14 $\pm$ 4.48                         \\
& \textbf{Aorta}          & \underline{ 4.14 $\pm$ 2.33}                          & 5.41 $\pm$ 3.32                          & \textbf{2.16 $\pm$ 1.22}                & \textbf{2.26 $\pm$ 0.75}                  & \underline{ 2.82 $\pm$ 1.33}  & 5.27 $\pm$ 4.90                         \\
\multirow{-5}{*}{\textbf{a-SSD(mm)}}       
& \textbf{Mean}           & \textbf{4.53}                                                            & \underline{ 5.94}              & 7.17             & \underline{3.24}                                      & \textbf{2.61}                           & 8.15                                    \\
\midrule[0.8pt]
& \textbf{Esophagus}      & \textbf{79.32 $\pm$ 7.98}                                       & \underline{ 77.58 $\pm$ 6.96}  & 74.22 $\pm$ 11.25                       & 76.66 $\pm$ 7.94                          & \underline{ 79.67 $\pm$ 6.67} & \textbf{82.10 $\pm$ 6.82}               \\
& \textbf{Heart}          & \textbf{95.73 $\pm$ 3.57}                                       & 92.84 $\pm$ 6.13                         & \underline{ 95.10 $\pm$ 4.60}   & \underline{ 96.34 $\pm$ 3.64}   & \textbf{96.60 $\pm$ 2.27}               & 92.70 $\pm$ 8.71                        \\
& \textbf{Trachea}        & 77.01 $\pm$ 8.83                                                & \textbf{85.75 $\pm$ 6.56}                & \underline{ 78.43 $\pm$ 6.78} & \underline{ 84.60 $\pm$ 7.82}    & \textbf{90.98 $\pm$ 6.73}               & 80.29 $\pm$ 10.06                       \\
& \textbf{Aorta}          & \underline{ 89.70 $\pm$ 3.38}                          & 87.79 $\pm$ 5.68                         & \textbf{90.51 $\pm$ 3.84}               & \textbf{92.28 $\pm$ 2.98}                 & \underline{ 91.76 $\pm$ 2.78} & 90.87 $\pm$ 3.10                        \\
\multirow{-5}{*}{\textbf{a-Preci(\%)}}        
& \textbf{Mean}           & \underline{ 85.44}                                    & \textbf{85.99}                           & 84.56                                   & \underline{ 87.47}              & \textbf{89.75}                          & 86.49
\\
\bottomrule[1.5pt]
\end{tabular*}
}
\label{tabblock}
\end{center}
\end{table*}


\subsubsection{Blocks'~Sensitivity}
Firstly, we assessed the effect of smoothing block configurations $\{k_l,k_m,k_r\}$, where $\{k_l,k_m,k_r\}$ denote the $k_l$, $k_m$, and $k_r$ smoothing iterations in the LSB, CSB, and RSB, respectively. Here, we first fixed the channel configuration with $p=32$ as the default. Table~\ref{tabblock} shows a quantitative results of different block parameter sets, and we observed from the pre-smoothing experiments (\emph{fixing} $k_m$ and $k_r$, \emph{varying} $k_l$) that 2D FAS-UNet with $k_l=3$ iterations achieved the best a-DSC score and precision of 85.60\% and 85.99\%, respectively. For~the post-smoothing (\emph{fixing} $k_l$ and $k_m$, \emph{varying} $k_r$), we saw that the model with $k_r=4$ iterations achieved the highest values of a-SSD and precision, the~a-DSC score being slightly lower than the model with the block configuration $\{3, 7, 5\}$ by 0.04\%. To~balance the prediction performance and computational costs, we set the block configuration as $\{3, 7, 4\}$ in all 2D~experiments.

\begin{table*}
\begin{center}
\caption{Quantitative assessment with a-DSC, a-SSD and a-Preci values of different initialization techniques (zero-initialization, random initialization and feature extraction initialization $\psi(\mathcal{K}^0(f)))$ using the proposed 2D FAS-Unet framework on the 2D SegTHOR validation datasets with four organs: esophagus, heart, trachea and aorta. "Mean" denotes an average score segmenting all organs. The best and second places are highlighted in Bold font and underlined ones, respectively.}
    \setlength{\tabcolsep}{0.8mm}{
\begin{tabular*}{\hsize}{@{\extracolsep{\fill}}clccc}
\toprule[1.5pt]
\multicolumn{2}{c}{\textbf{Feature Initialization}} & \textbf{Zero Initi} & 
\textbf{Random Initi} & $\psi(\mathcal{K}^0(f))$\textbf{-Initi} \\
\midrule[0.8pt]
\multicolumn{2}{c}{\textbf{Params}}     & 0.52M      & 0.52M         & 0.52M   \\
\midrule[0.8pt]
& \textbf{Esophagus} & 72.56 $\pm$ 11.66                        & \textbf{75.84 $\pm$ 10.02}                & \underline{ 74.24 $\pm$ 8.01}  \\
& \textbf{Heart}     & \underline{ 92.40 $\pm$ 4.48}   & 92.37 $\pm$ 5.06                          & \textbf{94.22 $\pm$ 1.52}                \\
& \textbf{Trachea}   & 84.21 $\pm$ 4.27                         & \textbf{85.08 $\pm$ 3.38}                 & \underline{ 84.47 $\pm$ 5.33}  \\
& \textbf{Aorta}     & \textbf{91.17 $\pm$ 4.49}                & \underline{ 89.81 $\pm$ 4.22}   & 89.39 $\pm$ 7.26                         \\
\multirow{-5}{*}{\textbf{a-DSC(\%)}} & \textbf{Mean}      & 85.08                                    & \textbf{85.77}                            & \underline{ 85.58}             \\
\midrule[0.8pt]
& \textbf{Esophagus}    & \textbf{2.39 $\pm$ 1.2}                 & 2.85 $\pm$ 0.87                        & \underline{ 2.67 $\pm$ 0.79} \\
& \textbf{Heart}        & \underline{ 15.38 $\pm$ 21.8} & 16.82 $\pm$ 32.15                      & \textbf{1.94 $\pm$ 0.66}               \\
& \textbf{Trachea}      & 5.67 $\pm$ 2.98                         & \underline{ 4.66 $\pm$ 3.59} & \textbf{3.01 $\pm$ 2.48}               \\
& \textbf{Aorta}        & \underline{ 2.85 $\pm$ 1.97}  & 7.59 $\pm$ 8.94                        & \textbf{2.82 $\pm$ 1.33}               \\
\multirow{-5}{*}{\textbf{a-SSD(mm)}}    & \textbf{Mean}         & \underline{ 6.57}             & 7.98                                   & \textbf{2.61}                          \\
 \midrule[0.8pt]
 & \textbf{Esophagus} & \textbf{81.8 $\pm$ 6.41}                 & \underline{ 81.3 $\pm$ 6.49}    & 79.67 $\pm$ 6.67                        \\
 & \textbf{Heart}     & 92.53 $\pm$ 8.43                         & \underline{ 93.31 $\pm$ 8.25}   & \textbf{96.6 $\pm$ 2.27}                \\
& \textbf{Trachea}   & 79.92 $\pm$ 8.34                         & \underline{ 79.0 $\pm$ 5.46}    & \textbf{90.98 $\pm$ 6.73}               \\
& \textbf{Aorta}  & \textbf{91.92 $\pm$ 2.75}                & 88.57 $\pm$ 5.28                          & \underline{ 91.76 $\pm$ 2.78} \\
\multirow{-5}{*}{\textbf{a-Preci(\%)}}  & \textbf{Mean}  & \underline{ 86.55}             & 85.55                                     & \textbf{89.75}                       \\
\bottomrule[1.5pt]
\end{tabular*}
}
\label{tabinit}
\end{center}
\end{table*}

\subsubsection{The Input Feature~Initialization}
We also evaluated the initialization configuration of the input feature $\bm{b}^0=\mathcal{K}^0f$ on the finest cycle $\ell=1$ to verify its sensitivity.
We compared different variants of the initialization method, such as zero initialization, random normal distribution initialization, and $\psi(\mathcal{K}^{0}f)$ initialization with the batch normalization operation $\psi(\cdot)$, where $\mathcal{K}^{0}$ was obtained by learning the convolutional kernel with a size of $p\times 3\times 3$ for 2D segmentation or~$p\times 3\times 3\times 3$ for 3D~segmentation.

Table~\ref{tabinit} shows the quantitative comparison of these variants. The~model with $\psi(\mathcal{K}^{0}f)$ initialization achieved a-DSC, a-SSD, and a-Preci values of 85.58\% (ranked second), 2.61~mm (top-ranked), and 89.75\% (top-ranked), respectively. Although~the model with random initialization had the highest a-DSC score, the~a-SSD and a-Preci scores were significantly lower than the model with $\psi(\mathcal{K}^{0}f)$ initialization. To~this end, we set the proposed
framework with $\psi(\mathcal{K}^{0}f)$ initialization as the default in this~work.

\subsubsection{Weight~Sharing}
To demonstrate the flexibility of the proposed framework, which does not have to be the
different $\mathcal{K}^\prime_{q,\ell,j}$ parameter configurations in different nonlinear $\mathcal{F}_{\mathcal{K}^\prime_{q,j}}^\ell$ (with respect to $j$) within the $\ell^{th}$ pre-smoothing ($q=l$) or post-smoothing ($q=r$) block, we
conducted several variants that had different
sharing settings among $\mathcal{K}^\prime_{q,\ell,j}$ (with respect to the iteration step~$j$).

We only varied the weight sharing settings to verify the sensitivity of the model. We compared four variants, and~the results are shown in Table~\ref{tabshare}. From~the evaluation metrics, we see that the model without the weight sharing configuration had more
 parameters and achieved the best performance on the a-DSC, a-SSD, and a-Preci values, respectively. The~performance of the other three models did not show a significant differences. Therefore, we adopted the default unshared
version in the rest of this~paper.

\begin{table*}
\begin{center}
\caption{Quantitative assessment with a-DSC, a-SSD and a-Preci values of using the proposed 2D FAS-Unet framework with/without weight-sharing of pre-/post smoothing steps 
 on the 2D SegTHOR validation datasets with four organs: esophagus, heart, trachea and aorta. "Mean" denotes an average score segmenting all organs. The best and second places are highlighted in Bold font and underlined ones, respectively.
}
    \setlength{\tabcolsep}{0.8mm}{
\begin{tabular*}{\hsize}{@{\extracolsep{\fill}}clcccc}
\toprule[1.5pt]
\multicolumn{2}{c}{\textbf{Block Weight Sharing}} & \_  & \textbf{LSB}& \textbf{RSB}& \textbf{LSB and RSB}                       \\
\midrule[0.8pt]
\multicolumn{2}{c}{\textbf{Params}}          & 0.52M                                  & 0.45M                                    & \underline{ 0.41M}             & \textbf{0.34M}                            \\
\midrule[0.8pt]
  & \textbf{Esophagus} & \textbf{74.24 $\pm$ 8.01}              & \underline{ 72.53 $\pm$ 10.17} & 73.01 $\pm$ 9.91                         & 72.53 $\pm$ 10.28                         \\
& \textbf{Heart}     & \textbf{94.22 $\pm$ 1.52}              & 90.51 $\pm$ 6.98                         & 89.33 $\pm$ 5.11                         & \underline{ 92.00 $\pm$ 3.97}    \\
& \textbf{Trachea}   & \textbf{84.47 $\pm$ 5.33}              & 84.14 $\pm$ 5.21                         & \underline{ 84.55 $\pm$ 2.60}   & 83.37 $\pm$ 4.10                           \\
& \textbf{Aorta}     & 89.39 $\pm$ 7.26                       & \textbf{90.46 $\pm$ 2.99}                & \underline{ 90.17 $\pm$ 4.30}   & 90.03 $\pm$ 2.40                           \\
\multirow{-5}{*}{\textbf{a-DSC(\%)}}  & \textbf{Mean}      & \textbf{85.58}                         & 84.41                                    & 84.26                                    & \underline{ 84.48}              \\
\midrule[0.8pt]
 & \textbf{Esophagus}  & \textbf{2.67 $\pm$ 0.79}  & 3.22 $\pm$ 0.97   & \underline{ 2.69 $\pm$ 0.72} & 4.26 $\pm$ 1.91                          \\
& \textbf{Heart}      & \textbf{1.94 $\pm$ 0.66}  & 22.37 $\pm$ 16.99 & 34.91 $\pm$ 31.76                      & \underline{ 14.51 $\pm$ 19.82} \\
& \textbf{Trachea}    & \textbf{3.01 $\pm$ 2.48}  & 6.30 $\pm$ 4.50     & \underline{ 5.17 $\pm$ 3.36} & 5.96 $\pm$ 3.38                          \\
& \textbf{Aorta}      & \textbf{2.82 $\pm$ 1.33}  & 4.64 $\pm$ 3.87   & \underline{ 4.61 $\pm$ 1.78} & 8.12 $\pm$ 5.49                          \\
\multirow{-5}{*}{\textbf{a-SSD(mm)}}   
& \textbf{Mean}       & \textbf{2.61}             & 9.13              & 11.85                                  & \underline{ 8.21}              \\
\midrule[0.8pt]
& \textbf{Esophagus} & \textbf{79.67 $\pm$ 6.67}              & \underline{ 77.94 $\pm$ 7.67}  & 77.36 $\pm$ 6.60                          & 74.64 $\pm$ 8.14                          \\
& \textbf{Heart}     & \textbf{96.60 $\pm$ 2.27}               & 88.05 $\pm$ 11.50                         & 88.72 $\pm$ 9.14                         & \underline{ 92.19 $\pm$ 7.80}    \\
& \textbf{Trachea}   & \textbf{90.98 $\pm$ 6.73}              & \underline{ 82.34 $\pm$ 9.26}  & 78.14 $\pm$ 5.00                          & 77.36 $\pm$ 8.26                          \\
& \textbf{Aorta}     & \textbf{91.76 $\pm$ 2.78}              & \underline{ 88.84 $\pm$ 3.57}  & 87.08 $\pm$ 6.09                         & 85.82 $\pm$ 3.77                          \\
\multirow{-5}{*}{\textbf{a-Preci(\%)}}   
& \textbf{Mean} & \textbf{89.75}                         & \underline{ 84.29}             & 82.82                                    & 82.50                                     \\
\bottomrule[1.5pt]
\end{tabular*}
}
\label{tabshare}
\end{center}
\end{table*}

\subsubsection{Effects of Varying the Number of~Channels}
In this section, we analyze the segmentation performance of the proposed method with varying the number $p$ of feature channels. In the~FAS-Solution module, the~pre-smoothing and post-smoothing steps (with $k_l$ and $k_r$ iterations, respectively) had the same update structure with the coarsest smoothing step, which included $k_m$ iterations. Therefore, we set the number of smoothing iterations as $\{k_l,k_m,k_r\}=\{3,7,4\}$ and~adopted a series of channel parameters $p=80, 64, 48, 32, 16$, respectively, for~comparison. Here, $p=80$ means that, in each convolutional operation of the FAS-Solution module, there were $80$ filters with the same kernel size of $3 \times 3$.

\begin{table*}
\begin{center}
\caption{Quantitative assessment with a-DSC, a-SSD and a-Preci values of different channel number $p$ using the proposed 2D FAS-Unet framework on the 2D SegTHOR validation datasets with four organs: esophagus, heart, trachea and aorta. "Mean" denotes an average score segmenting all organs. The best and second places are highlighted in Bold font and underlined ones, respectively.}
    \setlength{\tabcolsep}{0.6mm}{
\begin{tabular*}{\hsize}{@{\extracolsep{\fill}}clccccc}
\toprule[1.5pt]
\multicolumn{2}{c}{\textbf{Number of channels}} & \textbf{80}& \textbf{64}& \textbf{48} & \textbf{32}                                       & \textbf{16 }                                      \\
\midrule[0.8pt]
\multicolumn{2}{c}{\textbf{Params}}               & 3.24M                                      & 2.08M                                     & 1.17M                        & \underline{ 0.52M}             & \textbf{0.13M}                           \\
\midrule[0.8pt]
                            & \textbf{Esophagus}      & \underline{ 75.20 $\pm$ 11.53}    & \textbf{75.34 $\pm$ 12.59}                 & 74.50 $\pm$ 11.62             & 74.24 $\pm$ 8.01                         & 70.30 $\pm$ 11.04                         \\
                            & \textbf{Heart}          & \underline{ 94.19 $\pm$ 1.95}    & 93.79 $\pm$ 1.79                          & 92.04 $\pm$ 5.66             & \textbf{94.22 $\pm$ 1.52}                & 93.17 $\pm$ 2.41                         \\
                            & \textbf{Trachea}        & 84.77 $\pm$ 4.08     & \underline{ 86.97 $\pm$ 5.05}   & \textbf{87.29 $\pm$ 3.65}    & 84.47 $\pm$ 5.33                         & 85.86 $\pm$ 3.62                         \\
                            & \textbf{Aorta}    & \textbf{91.46 $\pm$ 5.14}    & \underline{ 91.20 $\pm$ 3.45 }   & 90.20 $\pm$ 6.76              & 89.39 $\pm$ 7.26                         & 90.17 $\pm$ 5.29                         \\
\multirow{-5}{*}{\textbf{a-DSC(\%)}}      & \textbf{Mean}           & \underline{86.41}           & \textbf{86.83}              & 86.01                        & 85.58                                    & 84.88                                    \\
\midrule[0.8pt]
 & \textbf{Esophagus}                     & 2.72 $\pm$   0.98           & \underline{ 2.49 $\pm$ 1.22}  & \textbf{2.37 $\pm$ 1.26} & 2.67 $\pm$ 0.79          & 3.18 $\pm$ 1.71                        \\
                                              & \textbf{Heart}                         & 8.33 $\pm$ 12.2             & \underline{ 8.45 $\pm$ 16.54} & 12.79 $\pm$ 26.80         & \textbf{1.94 $\pm$ 0.66} & 12.11 $\pm$ 18.35                      \\
                                              & \textbf{Trachea}                       & 7.52 $\pm$ 4.88             & 4.61 $\pm$ 3.50                         & 3.66 $\pm$ 2.63          & \textbf{3.01 $\pm$ 2.48} & \underline{ 3.20 $\pm$ 1.95}  \\
                                              & \textbf{Aorta}                    & \textbf{1.99 $\pm$ 1.02}    & 5.20 $\pm$ 4.92                          & 2.80 $\pm$ 1.73           & 2.82 $\pm$ 1.33          & \underline{ 2.38 $\pm$ 1.33} \\
\multirow{-5}{*}{\textbf{a-SSD(mm)}}                    & \textbf{Mean}                          & \underline{ 5.14} & 5.19                                    & 5.40                      & \textbf{2.61}            & 5.22                                   \\
\midrule[0.8pt]
                           & \textbf{Esophagus} & \underline{81.02 $\pm$ 6.21}              & \textbf{81.19$\pm$10.70}                         &  80.33 $\pm$ 8.23 & 79.67 $\pm$ 6.67                         & 77.39 $\pm$ 8.61                        \\
                           & \textbf{Heart}     & 95.84 $\pm$ 3.90 & \underline{95.85 $\pm$ 3.45}                          & 91.52 $\pm$ 8.89                        & \textbf{96.60 $\pm$ 2.27}                 & 94.31 $\pm$ 2.92                        \\
                           & \textbf{Trachea}   & 80.30 $\pm$ 8.44        &  84.64 $\pm$ 9.50                      & \underline{ 86.50 $\pm$ 8.41 } & \textbf{90.98 $\pm$ 6.73}                & 85.14 $\pm$ 9.38                        \\
                           & Aorta     & \textbf{93.04 $\pm$ 2.36}              & 91.17 $\pm$ 3.81                           & \underline{ 91.93 $\pm$ 3.61} & 91.76 $\pm$ 2.78                         & 89.85 $\pm$ 3.56                        \\
\multirow{-5}{*}{\textbf{a-Preci(\%)}}  & \textbf{Mean}      & 87.55      & \underline{ 88.21  }                                 & 87.57           & \textbf{89.75}                           & 86.67
                                   \\
\bottomrule[1.5pt]
\end{tabular*}
}
\label{tabchannel}
\end{center}
\end{table*}

Table~\ref{tabchannel} shows the quantitative comparison of different $p$-configurations.
It reveals that, as~the number of channels increased, the~parameter of our model squarely increased. Additionally, the~networks with the numbers of $64$ and $16$ achieved a-DSC scores of 86.83\% (ranking first) and 84.88\% (ranking lowest), respectively. When the number of channels was less than 64, increasing the number of channels could improve the a-DSC value, and one can see from Table~\ref{tabchannel} that the number of channels had a significant impact on the performance of the model.
Based on this observation, the~configuration with 64 channels is a
preferable setting to balance the segmentation performance
and computational costs, and~we fixed $p=64$ throughout all
the 2D~experiments.

\begin{table*}
\begin{center}
\caption{The hyperparameter configurations of the proposed FAS-Unet framework on different datasets.}
    \setlength{\tabcolsep}{0.6mm}{
\begin{tabular*}{\hsize}{@{\extracolsep{\fill}}ccccccc}
\toprule[1.5pt]
\textbf{Datasets}    & \begin{tabular}[c]{@{}c@{}}\textbf{Layer}\\      \textbf{number} $\mathcal{L}$\end{tabular} & \begin{tabular}[c]{@{}c@{}}\textbf{Feature} \\      \textbf{number} $p$\end{tabular} & \{$k_l, k_m, k_r$\} & \begin{tabular}[c]{@{}c@{}}\textbf{Weight sharing}\\ on $\mathcal{K}^\prime_{q,\ell,j}$  \end{tabular} & \begin{tabular}[c]{@{}c@{}}\textbf{Feature} \\\textbf{initialization} \end{tabular} & \textbf{Params} \\ \midrule[0.8pt]
\textbf{2D SegTHOR}    & 5   & 64   & \{3, 7, 4\} & No     & $\psi(\mathcal{K}^0(f)))$ & 2.08M  \\
\textbf{3D HVSMR 2016} & 4   & 32  & \{3, 6, 2\} & No   & $\psi(\mathcal{K}^0(f)))$ & 1.01M  \\
\textbf{3D CHAOS CT}   & 4   & 32   & \{3, 5, 2\} & No    & $\psi(\mathcal{K}^0(f)))$       & 1.00M  \\
\bottomrule[1.5pt]
\end{tabular*}
}
\label{concfas}
\end{center}
\end{table*}

To provide insights into the model hyperparameter configurations of the proposed 3D FAS-UNet version on the 3D HVSMR-2016 datasets and 3D CHAOS-CT datasets, we also carried out a series of ablation experiments to investigate the influence of two key design variables, the number of channels and the number of convolutional blocks. The~evaluation indicated that the network performed better with the configurations $p=32$, $\{k_l,k_m,k_r\}=\{3,6,2\}$ for the 3D HVSMR-2016 datasets and $\{3,5,2\}$ for the 3D
CHAOS-CT datasets as the default. Here, we do not detail these comparisons.

Finally, we illustrate the hyperparameter configurations of the proposed FAS-UNet on each dataset throughout all experiments, as shown in Table~\ref{concfas}.

\subsection{The 2D FAS-UNet for the SegTHOR~Datasets}
We evaluated the proposed network on the 2D SegTHOR datasets and~compared the visualizations and quantitative metrics with the existing
state-of-the-art segmentation methods, including
2D UNet {\cite{UNet}, CA-Net~\cite{canet2020}, CE-Net~\cite{gu2019ce-net}, CPFNet~\cite{cpfnet}, ERFNet~\cite{romera2018erfnet}, UNet++ \cite{zhou2018nest}, and LinkNet~\cite{linknet}}.

\begin{table*}
\begin{center}
\caption{Performance comparisons between the proposed 2D FAS-Unet and the popular networks using a-DSC, a-SSD and a-Preci values on the 2D SegTHOR validation datasets with four organs: esophagus, heart, trachea and aorta. "Mean" denotes an average score segmenting all organs. The best and second places are highlighted in Bold font and underlined ones, respectively.}
    \setlength{\tabcolsep}{0.8mm}{
\begin{tabular*}{\hsize}{@{\extracolsep{\fill}}clcccccccc}
\toprule[1.5pt]
\multicolumn{2}{c}{\textbf{Method}}      & \textbf{2D U-Net}  & \textbf{CA-Net} & \textbf{CE-Net}                   & \textbf{CPFNet}  & \textbf{ERFNet}  & \textbf{UNet++}   & \textbf{LinkNet}  & \textbf{2D FAS-UNet}                        \\
\multicolumn{2}{c}{}      & \textbf{\cite{UNet}}  & \textbf{\cite{canet2020}} & \textbf{\cite{gu2019ce-net}}                   & \textbf{\cite{cpfnet}}  & \textbf{\cite{romera2018erfnet}}  & \textbf{\cite{zhou2018nest}}   & \textbf{\cite{linknet}}                               &    \\
\midrule[0.8pt]
\multicolumn{2}{c}{\textbf{Params}}                                                                 & 17.26M                                  & 2.78M                                   & 29.00M                                          & 30.65M                                   & \textbf{2.06M}                             & 9.05M                                    & 21.79M                                 & \underline{ 2.08M}       \\
\midrule[0.8pt]
& \textbf{Esophagus} & 73.77±13.23                             & \textbf{76.26±8.72}                     & 64.17±10.78                              & 66.58±10.02         & 69.68±5.93               & 69.63±9.28                              & 65.19±9.18               & \underline{ 75.34 $\pm$   12.59} \\
& \textbf{Heart}     & 93.42±3.270                             & 93.61±2.45                              & \textbf{94.01±1.17}                      & 92.05±4.39          & 92.59±5.29               & 93.46±2.36                              & 93.19±9.18               & \underline{ 93.79 $\pm$ 1.79}    \\
& \textbf{Trachea}   & 86.71±2.69                              & 85.58±4.38                              & 85.12±4.37                               & \textbf{87.41±2.84} & 79.10±5.79               & 85.63±5.78                              & 86.16±4.35               & \underline{ 86.97 $\pm$ 5.05}    \\
& \textbf{Aorta}     & 90.42±3.45                              & \textbf{91.2±1.59}                      & 88.38±3.95                               & 88.92±4.98          & 90.08±4.06               & 90.61±2.66                              & 88.94±3.61               & \underline{ 91.2 $\pm$ 3.45}     \\
\multirow{-5}{*}{\textbf{a-DSC(\%)}}
& \textbf{Mean}      & 86.08                                   & \underline{ 86.66}            & 82.92                                    & 83.74               & 82.86                    & 84.83                                   & 83.37                    & \textbf{86.83}                             \\
\midrule[0.8pt]
& \textbf{Esophagus} & 2.78 $\pm$   1.36                       & 4.14 $\pm$   1.46                       & \underline{ 2.54 $\pm$   0.92} & 2.57 $\pm$   0.79   & 3.92 $\pm$   0.93        & 4.77 $\pm$ 1.33                         & 2.88 $\pm$   1.22        & \textbf{2.49 $\pm$   1.22}                 \\
& \textbf{Heart}     & 14.37 $\pm$ 22.78                       & \underline{ 4.87 $\pm$ 7.09}  & \textbf{4.57 $\pm$ 5.31}                 & 16.02 $\pm$ 22.26   & 7.76 $\pm$ 15.63         & 9.99 $\pm$ 11.17                        & 8.42 $\pm$ 17.02         & 8.45 $\pm$ 16.54                           \\
& \textbf{Trachea}   & 6.05 $\pm$ 6.14                         & 6.85 $\pm$ 4.87                         & \underline{ 3.81 $\pm$ 2.05}   & 4.46 $\pm$ 3.91     & 8.62 $\pm$ 4.26          & 5.84 $\pm$ 5.38                         & \textbf{2.57 $\pm$ 1.35} & 4.61 $\pm$ 3.50                            \\
& \textbf{Aorta}     & 6.36 $\pm$ 6.51                         & \underline{ 3.64 $\pm$ 1.68}  & 5.42 $\pm$ 1.62 & 7.51 $\pm$ 5.53     & \textbf{3.36 $\pm$ 2.31} & 4.41 $\pm$ 2.29                         & 5.85 $\pm$ 4.02          & 5.2 $\pm$ 4.92                             \\
\multirow{-5}{*}{\textbf{a-SSD(mm)}}
& \textbf{Mean}      & 7.39                                    & \underline{ 4.88}              & \textbf{4.08}             & 7.64                & 5.92                     & 6.25                                    & 4.93                   & 5.19                                       \\
\midrule[0.8pt]
& \textbf{Esophagus} & \textbf{82.82 $\pm$ 4.69}               & \underline{ 82.67 $\pm$ 5.46} & 79.14 $\pm$ 7.39                         & 73.92 $\pm$ 7.29    & 68.78 $\pm$ 7.88         & 79.83 $\pm$ 7.72                        & 74.1 $\pm$ 7.32          & 81.19 $\pm$ 10.70                          \\
 & \textbf{Heart}     & 93.47 $\pm$ 5.39                        & 94.91 $\pm$ 3.79                        & \underline{ 95.26 $\pm$ 2.66}  & 91.85 $\pm$ 8.39    & 92.61 $\pm$ 9.62         & 94.27 $\pm$ 4.58                        & 95.00 $\pm$ 4.71         & \textbf{95.85 $\pm$ 3.45}                  \\
& \textbf{Trachea}   & \underline{ 85.21 $\pm$ 5.91} & 81.72 $\pm$ 7.29                        & 84.81 $\pm$ 9.74                         & 84.48 $\pm$ 6.90    & 72.46 $\pm$ 9.71         & \textbf{87.59 $\pm$ 9.05}               & 83.88 $\pm$ 9.26         & 84.64 $\pm$ 9.50                           \\
& \textbf{Aorta}     & 90.01 $\pm$ 2.81                        & 88.84 $\pm$ 4.04                        & 86.10 $\pm$ 4.73                         & 88.84 $\pm$ 2.92    & 88.07 $\pm$ 3.35         & \underline{ 90.39 $\pm$ 3.57} & 88.91 $\pm$ 5.28         & \textbf{91.17 $\pm$ 3.81}                  \\
\multirow{-5}{*}{\textbf{a-Preci(\%)}}
& \textbf{Mean}      & 87.88                                   & 87.03                                   & 86.33                                    & 84.77               & 80.48                    & \underline{ 88.02}            & 85.47                    & \textbf{88.21} \\
\bottomrule[1.5pt]
\end{tabular*}
}
\label{tabseg}
   \end{center}
\end{table*}

\begin{figure*}
\centering
\includegraphics[width=\textwidth]{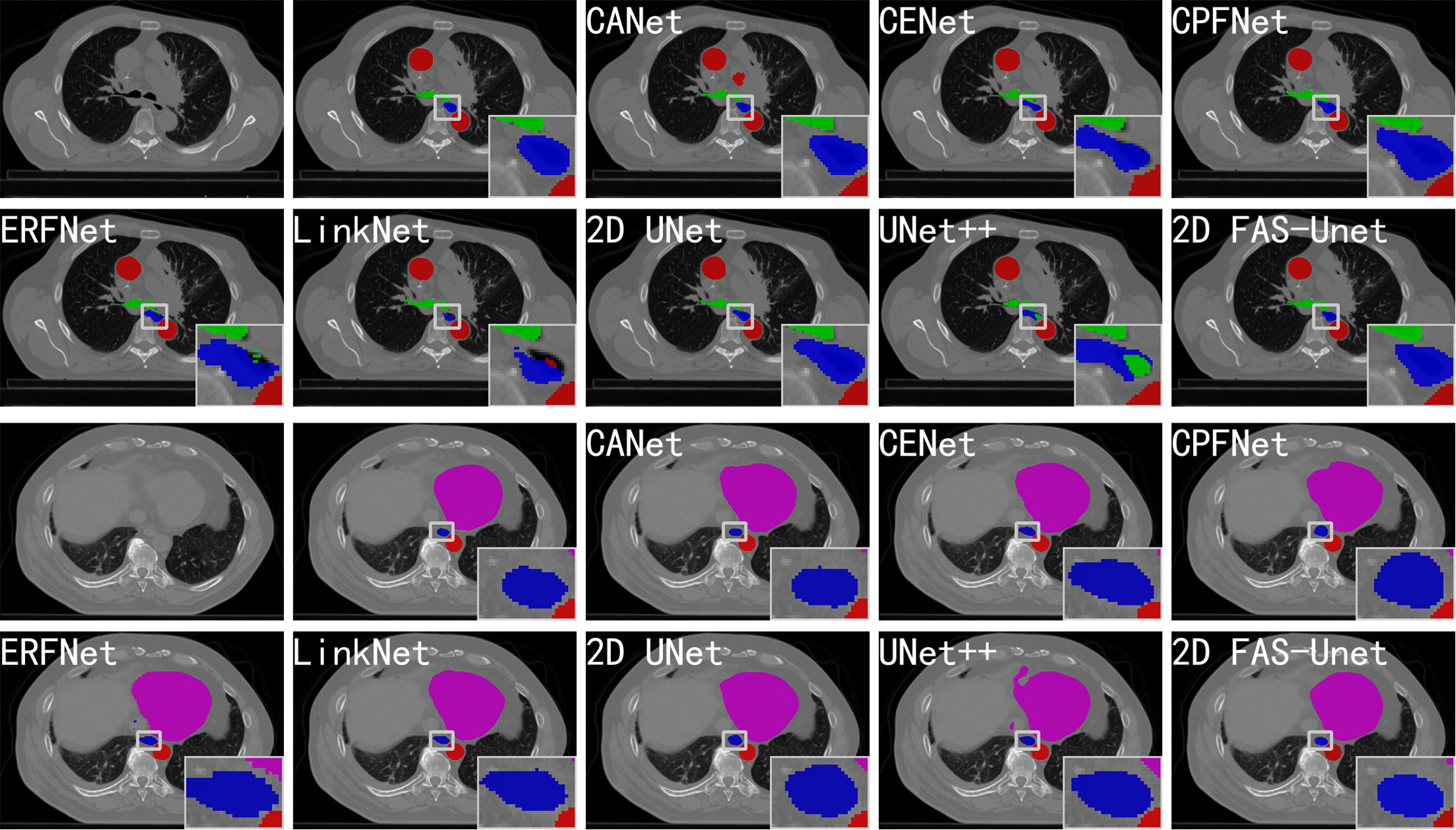}
\caption{\footnotesize {Comparison with the other state-of-the-art networks on validation set of 2D SegTHOR datasets. The blue, pink, green, and red regions represent the esophagus, heart, trachea and aorta, respectively. Form the left to right on first rows and third rows: the original images, ground truth, and the segmentation results of CANet, CE-Net, CPFNet, respectively. Form the left to right on second rows and fourth rows: the segmentation results of ERFNet, LinkNet, 2D U-Net, UNet++ and 2D FAS-Unet, respectively.}}
\label{segthorpred}
\end{figure*}

In Table~\ref{tabseg}, we show the quantitative results of each organ's segmentation compared with the other seven models. We can see that the segmentation performance of the heart was the best among all organs, and~its a-DSC score was more than 92\% for each method, followed by the aorta, and~the worst was the esophagus. The~main reason for the good performance in extracting the heart was that the heart region is the largest, and~its inner pixel value changes little, while its boundary is more obvious (see Figure~\ref{segthorpred}), while the esophageal region is the smallest among all organs, which increased the segmentation difficulty.

The proposed method achieved an a-DSC value of 86.83\% (ranked first) and an a-Preci value of 88.21\% (ranked first) with only 2.08 M parameters.
Compared with the state-of-the-art models CA-Net (second-ranked in the a-DSC score) and UNet++ (ranked second in the a-Preci value), the~proposed 2D FAS-UNet obtained a 0.17\% improvement of the a-DSC score with only 75\% as many parameters as CA-Net and a~0.19\% improvement of the a-Preci value with only 22\% as many parameters as UNet++. Our method also had a higher a-DSC score than the third-ranked UNet by 0.75\%, but~had far fewer parameters than the 17.26 M of UNet. Compared with ERFNet, which achieved a-DSC and a-Preci values of 82.86\% and 80.48\% with the fewest parameters, respectively, FAS-UNet was higher in the overall a-DSC rankings. The~a-DSC score of ERFNet ranked last, so we think it may fall into an under-fitting situation. This shows that ERFNet reduces the performance of the network while saving the parameters. CA-Net obtained a good a-DSC value because the attention mechanism may improve the segmentation results of small organs.

The evaluation results were also measured in terms of the a-SSD value for segmentation predictions of the eight methods. The~proposed method achieved an a-SSD value of 5.19mm (ranked fourth); CE-Net ranked first, which achieved an a-SSD value of 4.08mm;~ERFNet with the fewest parameters ranked fifth with 5.92mm. CA-Net and Link-net ranked second and third, which had 2.78 M and 21.79 M parameters, respectively. The~good results of CA-Net in terms of the a-SSD metric may be due to the use of multiple attention mechanisms, which enables the network to suppress the background region, and~the network has a stronger ability to recognize the object region. The~a-SSD score of UNet++ was much higher than that of 2D FAS-UNet, which shows that the over-segmented pixels were less than the under-segmented pixels. Meanwhile, we observed that the a-SSD value of our method was close to that of LinkNet in three organs; only the tracheal region was significantly worse than it, which makes it better than our method in the mean a-SSD score.

Figure~\ref{segthorpred} evaluates the visualizations of the segmented predictions obtained by the popular methods. One can observe that all methods except CA-Net (with an obvious over-segmentation) can accurately segment the aortic region (red).
The reason may be that, during the imaging process, the~aortic organ may be assigned to the very same image pixel, which
leads to a small difference of the internal pixel value; in~particular, the~network can accurately learn its features. Almost all methods can also approximately segment the trachea organ shape (green); only the organ boundary is not clearly visible. This may be a common problem in small object segmentation because the hard-to-detect small-scale feature will be degraded rapidly with convolution and pooling. All methods were able to extract the heart location (magenta), while the visual quality of the proposed method was significantly better than the state-of-the-art methods. Our approach achieved the least missing pixels
 at the organ boundary, which resulted in substantially better performance than the existing results; the~visualization remained comparable. It should also be emphasized that only a few methods performed well on the left boundary because the left side of the heart's boundary is very blurry. For~example, UNet++ presented over-segmentation, and~CA-Net on the left boundary was significantly different from the ground-truth, while the results of ERFNet and LinkNet had significant differences compared with the ground-truth in the shape aspect. This also shows that our method is robust to the heavy occlusion of illumination and large background clutters.

For the esophagus organ (blue) in Figure~\ref{segthorpred}, the~segmented results of CE-Net, CPFNet, and~UNet had significantly differences compared with the ground-truth in the shape aspect. The~results of ERFNet and UNet++ had some trachea pixels within the esophagus, and~LinkNet's prediction had some aortic pixels, all of which were clearly error-segmented. The~predictions of CA-Net and FAS-UNet were similar to manual segmentation, but~the result of CA-Net had a small aorta patch in the background region, while our method obviously performed better. The~reasons for the esophagus's bad results were the fuzzy boundary and the small pixel value difference; in~particular, the~esophagus and aorta almost overlap in the second slice. Therefore, the~proposed method is more robust, and it is more~difficult for it to be affected by noise. This further shows that our method is effective in medical image segmentation.

\subsection{The 3D FAS-UNet for the HVSMR-2016~Datasets}
We also conducted the segmentation experiments of our 3D FAS-UNet on the HVSMR-2016 datasets. We compared our predictions with seven baseline models including 3D UNet~\cite{3DUnet}, AnatomyNet~\cite{2019AnatomyNet}, DMFNet~\cite{chen2019dmfnet}, HDC-Net~\cite{luo2020hdc}, RSANet~\cite{zhang2019rsanet}, Bui-Net~\cite{bui2019skip-connected}, and VoxResNet~\cite{Chen2018VoxResNetDV}.
Table~\ref{tabhvs} shows the segmentation results of different methods. Clearly, our method with fewer parameters ranked second in both the a-DSC and a-SSD values and~obtained the top rank in mean precision.

\begin{table*}
    \begin{center}
    \caption{Performance comparisons between the proposed 3D FAS-Unet and the popular networks using a-DSC, a-SSD and a-Preci values on 3D HVSMR-2016 validation datasets with both organs: myocardium and blood pool. "Mean" denotes an average score segmenting all organs. The best and second places are highlighted in Bold font and underlined ones, respectively.}
    \setlength{\tabcolsep}{1mm}{
\begin{tabular*}{\hsize}{@{\extracolsep{\fill}}llcccccccc}
\toprule[1.5pt]
\multicolumn{2}{c}{\textbf{Method}}  & \textbf{3D U-Net}& \textbf{AnatomyNet}    & \textbf{DMFNet}& \textbf{HDC-Net} & \textbf{RSANet} & \textbf{Bui-Net}  & \textbf{VoxResNet}  & \textbf{3D FAS-Unet}                                                 \\
\multicolumn{2}{c}{}  & \textbf{\cite{3DUnet}}  & \textbf{\cite{2019AnatomyNet}}             &\textbf{\cite{chen2019dmfnet}}  & \textbf{\cite{luo2020hdc}}    & \textbf{\cite{zhang2019rsanet}}  & \textbf{\cite{bui2019skip-connected}}  &  \textbf{\cite{Chen2018VoxResNetDV}}   &                                                  \\
\midrule[0.8pt]
\multicolumn{2}{c}{\textbf{Params}}                                                                                         & 8.16M                                     & \underline{ 0.73M}          & 3.87M                                         & \textbf{0.30M}                            & 24.55M                               & 2.53M                                            & 1.70M                                   &      1.01M    \\
\midrule[0.8pt]
& \textbf{MY}                     & \textbf{76.42 $\pm$   3.32}               & 69.50 $\pm$   4.69                    & 70.30 $\pm$ 1.03                               & 69.77 $\pm$   5.10                       & 71.94 $\pm$   0.69                   & 69.84 $\pm$   0.47                               & 65.79 $\pm$   6.00                       & \underline{76.42 $\pm$ 4.38}                   \\
 & \textbf{BP}                     &  89.21 $\pm$ 1.61  & 86.01 $\pm$ 1.51                     & \textbf{89.45 $\pm$ 0.18}                             & 88.03 $\pm$ 2.24                        & 84.71 $\pm$ 0.93                     & \underline{ 89.4 $\pm$ 0.46}                         & 84.66 $\pm$ 1.67                        & 89.01 $\pm$ 0.18                                            \\
\multirow{-3}{*}{\begin{tabular}[c]{@{}c@{}}\textbf{a-DSC (\%)}\end{tabular}}
& \textbf{Mean}                           & \textbf{82.82}                            & 77.76                                & 79.88                                         & 78.90                                   & 78.33                                & 79.62                                            & 75.23                                   & \underline{ 82.72}                                \\
\midrule[0.8pt]
                            & \textbf{MY} & \textbf{2.06 $\pm$   0.73}  & 4.03 $\pm$   0.31                       & 2.74 $\pm$   0.20         & 3.36 $\pm$   0.32   & \underline{2.31 $\pm$   0.30}  & 2.56 $\pm$   0.22                         & 2.52 $\pm$   1.06  &  2.32 $\pm$   0.27  \\
                           & \textbf{BP} & 3.05 $\pm$ 1.53             & 2.98 $\pm$ 1.21                         & \textbf{2.07 $\pm$ 0.48}  & 2.65 $\pm$ 0.02     & 4.01 $\pm$ 0.55    & 2.58 $\pm$ 0.51                           & 3.47 $\pm$ 0.04    & \underline{ 2.57 $\pm$ 1.15}    \\
\multirow{-3}{*}{\textbf{a-SSD(mm)}}
& \textbf{Mean}       & 2.56                        & 3.50                                    & \textbf{2.40}             & 3.00                & 3.16               & 2.57                                      & 2.99               & \underline{ 2.44}               \\
\midrule[0.8pt]
& \textbf{MY}                     & 81.02 $\pm$   9.13                        & 71.25 $\pm$   6.22                   & 80.27 $\pm$   4.22                            & 67.41 $\pm$   12.94                     & 76.76 $\pm$ 9.00                      & \underline{ 83.94 $\pm$   6.73}        & 77.94 $\pm$   10.40                      & \textbf{84.61 $\pm$   7.14}                                 \\
& \multicolumn{1}{l}{\textbf{BP}} & \multicolumn{1}{c}{87.44 $\pm$ 2.62}      & \multicolumn{1}{c}{89.73 $\pm$ 4.61} & \multicolumn{1}{c}{\textbf{91.78 $\pm$ 0.54}} & \multicolumn{1}{c}{87.84 $\pm$ 2.09}    & \multicolumn{1}{c}{82.86 $\pm$ 6.08} & \multicolumn{1}{l}{89.66 $\pm$ 0.56}             & \multicolumn{1}{c}{84.29 $\pm$ 1.13}    & \multicolumn{1}{c}{\underline{ 91.18 $\pm$ 2.36}} \\
\multirow{-3}{*}{\textbf{a-Preci (\%)}}                                                        & \multicolumn{1}{l}{\textbf{Mean}}       & \multicolumn{1}{c}{84.23}                 & \multicolumn{1}{c}{80.49}            & \multicolumn{1}{c}{86.03}                     & \multicolumn{1}{c}{77.62}               & \multicolumn{1}{c}{79.81}            & \multicolumn{1}{c}{\underline{ 86.80}} & \multicolumn{1}{l}{81.11}               & \multicolumn{1}{c}{\textbf{87.90}}                  \\
\bottomrule[1.5pt]
\end{tabular*}
}\label{tabhvs}
  \end{center}
\end{table*}

The proposed method achieved an a-DSC value of 82.75\% (ranked second) with only 1.01 M parameters and~followed the first-ranked 3D UNet by 0.1\% in the a-DSC score with only 15\% as many parameters as 3D UNet. Meanwhile, our method outperformed the third-ranked DMFNet by 2.84\%. Although~the numbers of parameters of HDC-Net and AnatomyNet were lower than that of our method, the~a-DSC score of 3D FAS-UNet was 3.82\% and 4.96\% higher than theirs, respectively. One may notice that a black-box (unexplainable) network with a small number of parameters has low segmentation performance, which may be due to under-fitting. However, the~number of parameters in RSANet is a bit large, and the~effect was also not good enough. This may be due to too little training data, so the model appears to be over-fitting.
Our method also obtained an a-SSD value of 2.44 mm (2nd rank), which was lower than DMFNet's 2.40 mm (1st rank) by 0.12 mm, and~it slightly improved compared with 3D UNet, whose a-SSD value was 2.56 mm (3rd ranked). Although~3D FAS-UNet had a slightly lower a-SSD than DMFNet, it had 2.86 M fewer parameters. Compared with HDC-Net and AnatomyNet with fewer parameters, our method performed better on the a-SSD metric for myocardium and blood pool. The~result of the a-SSD value shows that our method has good performance in segmenting object boundaries. The~proposed method obtained the best a-Preci score of 87.90\%, which was higher than the second-ranked Bui-Net by 1.10\%. Although~3D FAS-UNet ranked third in the number of parameters, all three metrics were better than HDC-Net and AnatomyNet with fewer parameters. Therefore, our method achieves a good balance between the number of parameters and performance. Experiments on these datasets showed that the proposed FAS-driven explainable model can be robustly applied to 3D medical segmentation tasks.

\begin{figure*}
\centering
\includegraphics[width=0.95\linewidth]{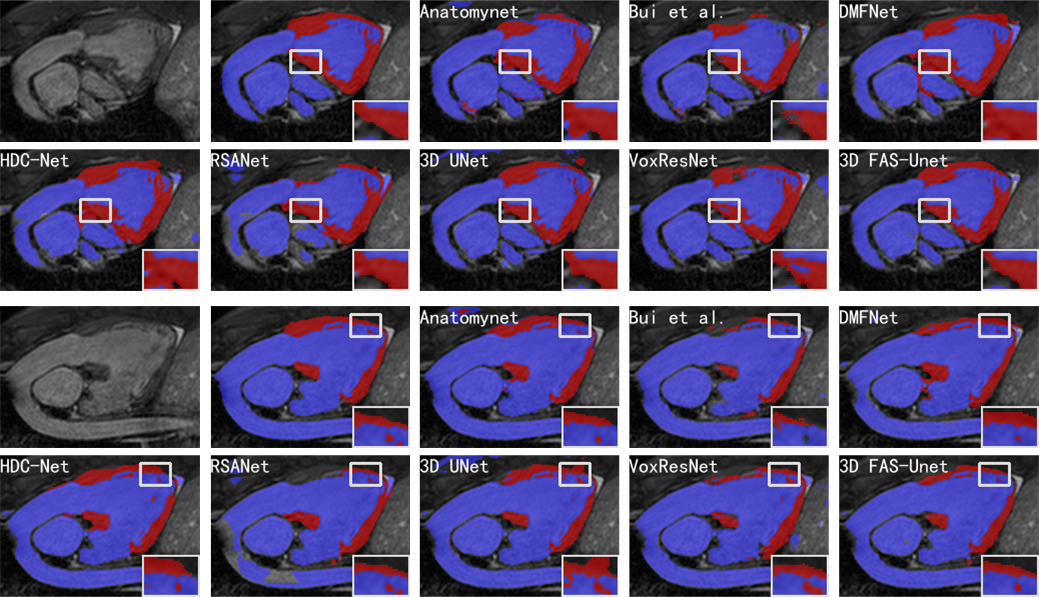}
\caption{Visualizations of different methods for cardiovascular MR segmentation of different slices. From the left to right on first and third rows: the original images, ground truth, and the segmentation results of AnatomyNet, Skip-connected 3D DenseNet, DMFNet, respectively. From the left to right on second and fourth rows: the segmentation results of HDC-Net, RSANet, 3D U-Net, VoxResNet, and 3D FAS-UNet, respectively. The blue and red colors represent blood pool and myocardium, respectively.}
\label{hvsmr}
\end{figure*}

Figure~\ref{hvsmr} visualizes the segmentation results of different methods on two slices of the HVSMR-2016 datasets, and it can be clearly observed that the proposed model highlighted less over-segmented regions outside the ground-truth compared with other methods. Meanwhile, it also can show that it was hard for our method to be affected by the voxels in the background region, where it did not predict the voxels of the background region as blood pools or myocardium, but most other methods predicted more background voxels as the object. The~results showed that these methods are easily affected by noise in the background region.

In general, one can observe that AnatomyNet, VoxResNet, and~3D UNet showed obvious segmentation noise (over-segmented region). The~reason is that the network collects much noise information in the interactions from input data of the network due to a too simple data pre-processing method, which affects the feature extraction. Several methods presented over-segmentation in the myocardial zoom-in, because~the pixel value of this organ is very close to the background. The~myocardium is structurally distorted, which makes the shape of the myocardium completely different compared to normal/healthy myocardium. Although~VoxResNet did not have this phenomenon, it divided the middle part of the myocardial region into blood pools, which was also an obvious error segmentation. Only 3D UNet and our method performed better; especially, our method was closer to the ground-truth in shape. Further, all methods had poor segmentation results in the upper myocardial region; the~intensity homogeneity between this organ and the upper background indicates that this region is very difficult to segment. We can see from the zoom-in results that many methods have obvious over-segmentation or under-segmentation for the myocardium and blood pool. Compared with Bui-Net and RSANet, we observed that the proposed learnable specialized FAS-UNet network still had obvious advantages in this region, and~the results were very close to the ground-truth in the myocardial region (red) with respect to the shape and size. For~the blood pool region (blue), our results did not show significant differences with other methods.

The proposed network integrates medical image data and the variational convexity MS model and algorithm (FAS scheme), and~implements them through convolution-based deep learning, so it may be possible to design specialized modules that automatically satisfy some of the physical invariants for better accuracy and robustness. Qualitative and quantitative experimental results demonstrated the effectiveness and superiority of our method. It can not only correctly locate the position of the myocardium, but~also segment the myocardium and blood pool in the complex marginal region. Moreover, the~integrity and continuity of our method in the object were well preserved. Overall, it performed better than the existing state-of-the-art methods in 3D medical image segmentation.

\subsection{The 3D FAS-UNet for the CHAOS-CT~Datasets}
In this part, {the proposed 3D FAS-UNet was compared with seven baseline models on the 3D CHAOS-CT datasets}, including 3D UNet~\cite{3DUnet}, Bui-Net~\cite{bui2019skip-connected}, DMFNet~\cite{chen2019dmfnet}, 3D ESPNet~\cite{mehta20193despnet}, RSANet~\cite{zhang2019rsanet}, RatLesNetV2~\cite{valverde2020ratlesnetv2}, and~HDC-Net~\cite{luo2020hdc}.

Firstly, we used a post-processing technique to improve the prediction results, where small undesirable clusters of voxels separated from the largest connection component may be over-segmented or the ``holes'' inside the liver may also be under-segmented. {Table \ref{tabliver}} shows the prediction results of different methods. Before~post-processing, the~proposed method achieved an a-DSC of 96.69\% (top-ranked) with only 1.00 M parameters, which is slightly higher than the second-ranked RSANet by 0.08\% in the mean a-DSC score with only about 4\% as many parameters as RSANet, and~further outperformed the third-ranked DMFNet by 0.28\%. Our method also obtained an a-SSD value of 4.04mm (ranked second), which is the same as RSANet (top-ranked). Although~3D FAS-UNet had a lower mean a-Preci value than 3D ESPNet by 1.54\%, it had 2.57 M fewer parameters.

The a-DSC scores of all methods were significantly improved by post-processing techniques. The~3D FAS-UNet achieved an a-DSC score of 97.11\% (ranked second) and~followed the first-ranked 3D UNet by 0.05\% in the mean a-DSC with only 15\% as many parameters as 3D UNet. Our method also obtained an a-SSD value of 1.16 mm (top-ranked), and~it was better than 3D UNet (ranked 2nd) and RSANet (ranked 3rd) by 0.07 mm and 0.10 mm, respectively. The 3D FAS-UNet achieved an a-Preci of 96.73\%, which is lower than 3D ESPNet and RatLesNetV2. Although~3D FAS-UNet ranked third in the parameter evaluation, all five metrics were better than HDC-Net and RatLesNetV2 with fewer parameters; only the a-Preci value with post-processing was {lower} than that of RatLesNetV2. Thus, our method achieves a good balance between the number of parameters and segmentation performance.

\begin{table*}
\begin{center}
\caption{Performance comparisons of liver segmentation between the different 3D networks with/without post-processing using a-DSC, a-SSD and a-Preci values on 3D CHAOS-CT validation datasets. The best and second places are highlighted in Bold font and underlined ones, respectively.}
    \setlength{\tabcolsep}{1mm}{
\begin{tabular*}{\hsize}{@{\extracolsep{\fill}}lccccccc}
\toprule[1.5pt]
    \multirow{2}{*}{\textbf{Method}} &
    \multirow{2}{*}{\textbf{Params}} &
    \multicolumn{3}{c}{\textbf{Without post-processing}} &
    \multicolumn{3}{c}{\textbf{With post-processing}} \\
\cmidrule[0.7pt]{3-5} \cmidrule[0.7pt](lr){6-8}          &       & \textbf{a-DSC(\%)} & \textbf{a-SSD(mm)} & \textbf{a-Preci(\%)} & {\textbf{a-DSC (\%)}} & { \textbf{a-SSD(mm)}} & {\textbf{a-Preci(\%)}}  \\
\midrule[0.8pt]
\textbf{3D U-Net} \cite{3DUnet}    & 8.16M       & 96.18 $\pm$ 1.72  & 5.65 $\pm$ 5.15  & 94.87 $\pm$ 4.12                        & {\textbf {97.16 $\pm$ 0.54}}    &  \underline{ 1.23 $\pm$ 0.28 }     & 96.71 $\pm$ 2.01                        \\
\textbf{Bui-Net} \cite{bui2019skip-connected}  & 2.24M  & 89.41 $\pm$ 4.86   & 11.54 $\pm$ 4.52 & 85.90 $\pm$ 8.20                          & 92.50 $\pm$ 4.02                         &  4.42 $\pm$ 3.60    & 91.85 $\pm$ 7.92
\\
\textbf{DMFNet} \cite{chen2019dmfnet}     & 3.87M     & 96.41 $\pm$ 0.99   & 4.45 $\pm$ 2.98 & 95.34 $\pm$ 3.28                        & 96.93 $\pm$ 0.47   & 1.29 $\pm$ 0.11 & 96.31 $\pm$ 2.42
\\
\textbf{3D ESPNet} \cite{mehta20193despnet}   &  3.57M                    & 95.72 $\pm$ 0.81                        & 4.32 $\pm$ 1.74                        & \textbf{97.46 $\pm$ 2.15}               & 96.15 $\pm$ 0.80                        & 1.68 $\pm$ 0.67                        & \textbf{98.33 $\pm$ 1.31}               \\
\textbf{RSANet} \cite{zhang2019rsanet}     & 24.54M                   & \underline{ 96.61 $\pm$ 1.00} & \textbf{4.04 $\pm$ 2.30}               & 95.23 $\pm$ 2.70                        & 97.08 $\pm$ 0.67                        & 1.26 $\pm$ 0.33                        & 96.11 $\pm$ 1.95                        \\
\textbf{RatLesNetV2} \cite{valverde2020ratlesnetv2} & \underline{0.83M  }                  & 95.73 $\pm$ 1.22                        & 7.04 $\pm$ 3.12                        & 95.13 $\pm$ 3.40                        & 96.82 $\pm$ 0.69                        & 1.27 $\pm$ 0.26                        & \underline{ 97.23 $\pm$ 2.12} \\
\textbf{HDC-Net} \cite{luo2020hdc}     & \textbf{0.29M }                   & 95.30 $\pm$ 1.64                        & 7.50 $\pm$ 4.92                        & 93.77 $\pm$ 4.23                        & 96.42 $\pm$ 0.67                        & 1.30 $\pm$ 0.17                        & 95.87 $\pm$ 2.51                        \\
\textbf{3D FAS-Unet} & 1.00M                    & \textbf{96.69 $\pm$ 0.76}               & \underline{ 4.04 $\pm$ 3.11} & \underline{ 95.92 $\pm$ 2.78} & \underline{ 97.11 $\pm$ 0.32} & \textbf{1.16 $\pm$ 0.13}               & 96.73 $\pm$ 2.08                        \\
    \bottomrule[1.5pt]
    \end{tabular*}
    }
\label{tabliver}
\end{center}
\end{table*}

\begin{figure*}
\centering
\includegraphics[width=0.95\linewidth]{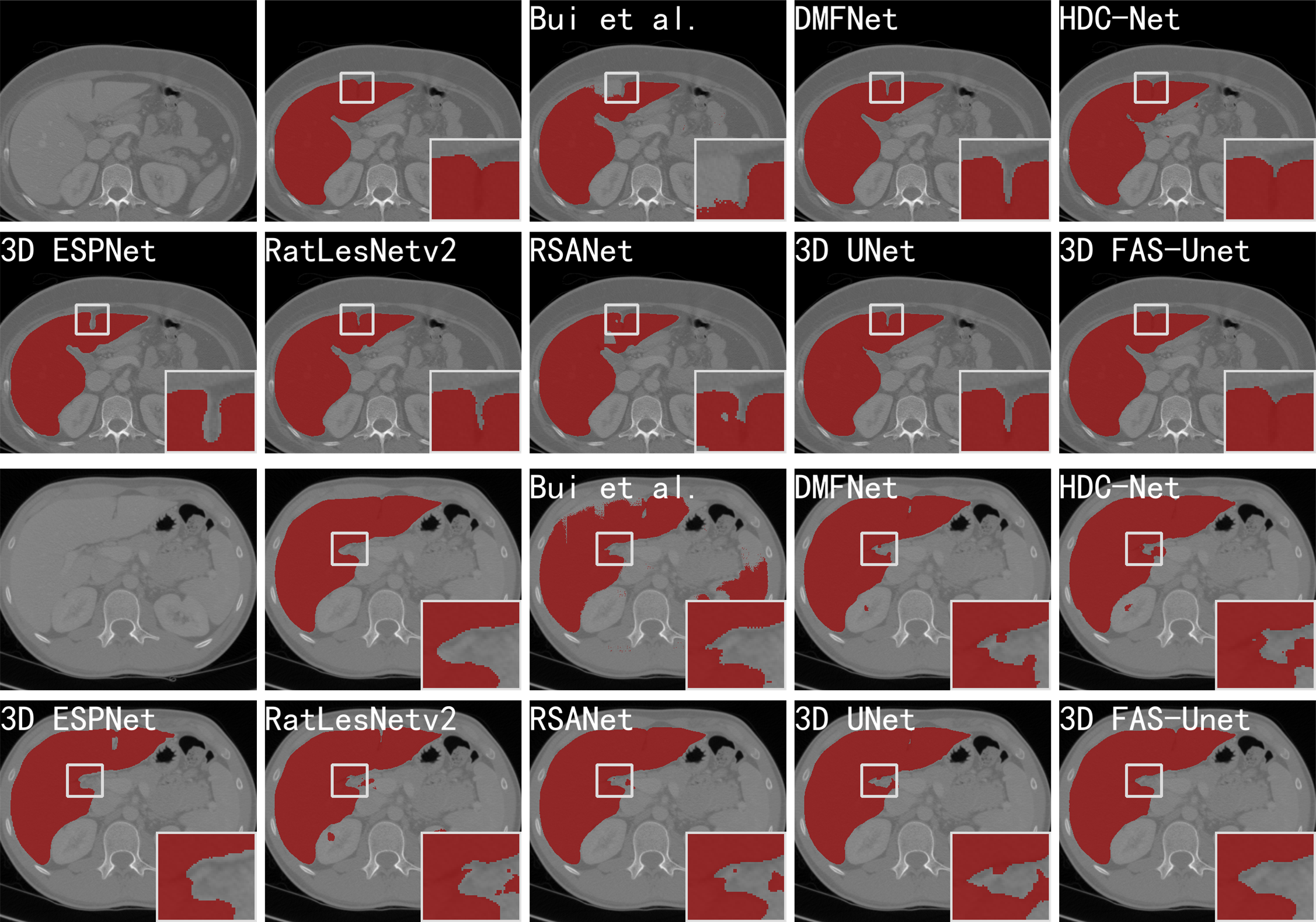}
\caption{Visualizations (without post-processed) of different methods for liver CT segmentation of two slices. From the left to right on  first and third rows: the original images, ground truth, the segmentation results of Skip-connected 3D DenseNet, DMFNet, HDC-Net, respectively. From the left to right on second and fourth rows: the segmentation results of 3D ESPNet, RatLesNetv2, RSANet, 3D UNet, and 3D FAS-Unet, respectively.}
\label{chaos}
\end{figure*}

Figure~\ref{chaos} visualizes the prediction results of different networks on two slices of the CHAOS-CT datasets, and it can be clearly observed that the other networks highlighted more over-segmented regions outside the {ground-truth} compared with our network.
Further, we can observe that the results for Bui-Net, DMFNet, HDC-Net, RatLesNetv2, and~RSANet showed obvious segmentation noise in the background region. Although~3D ESP-Net did not present this phenomenon, the~result showed an obvious ``hole'' inside of the liver, which was {an} obvious error-segmentation. However, our approach did not show these evident inaccurate results. In~addition, most methods showed over-segmentation or under-segmentation on the boundaries of liver because it is very blurred in the CT image.
From the zoom-in results of the first two rows of Figure~\ref{chaos}, we can see that all mentioned methods had obvious under-segmentation except HDC-Net and our method, but~our method had less noise in the background region.
From the zoom-in results of the last two rows in Figure~\ref{chaos}, we observe that many methods showed obvious over-segmentation on the liver boundaries. Only our method and 3D ESPNet achieved a good performance in this region, but~3D ESPNet extracted a ``hole'' in the liver region. In~summary, compared to other methods, the~boundary results obtained by our method were smoother, and~the shape of the liver was more similar to the ground-truth.

\begin{figure}
  \centering
  \includegraphics[width=1.0\linewidth]{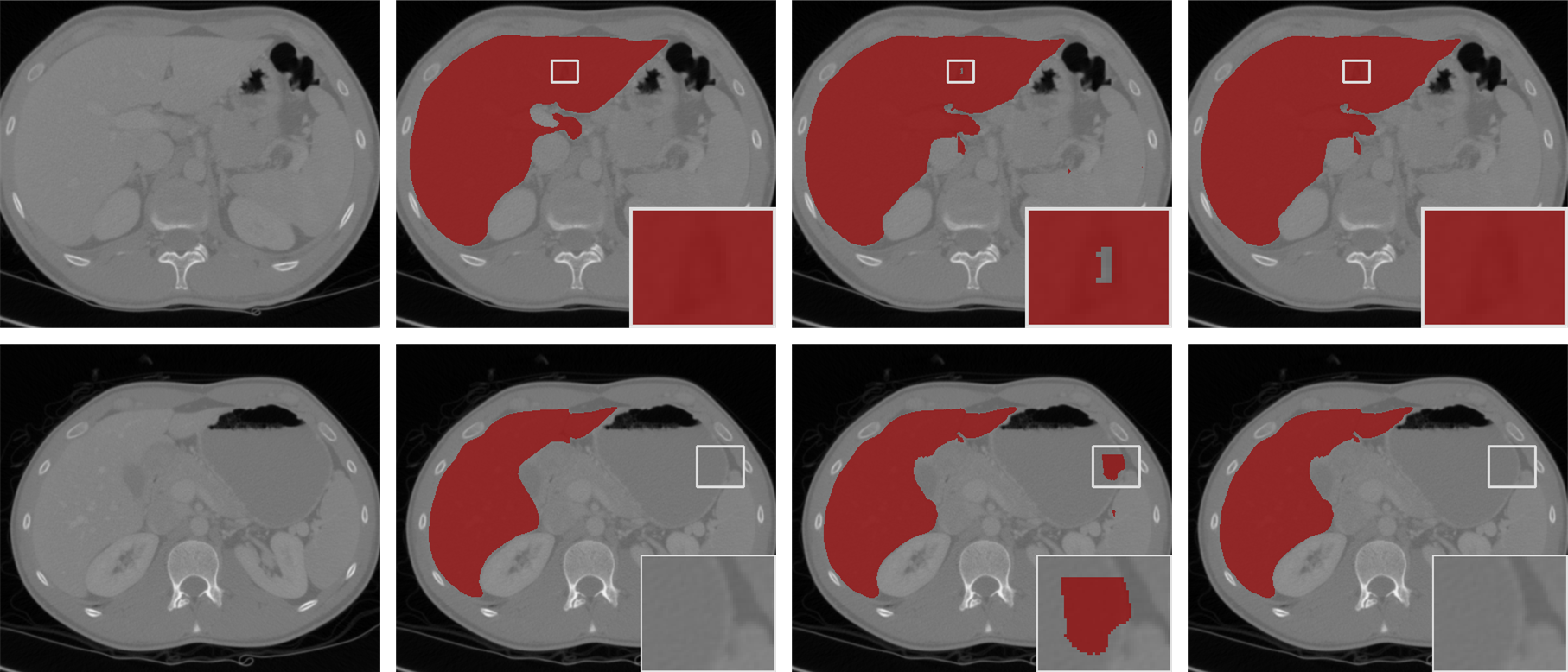}
  \caption{The segmentation results with post-processing. From left to right: the selected scans of the validation set (first column), ground truths (second column), segmentation results of 3D FAS-Unet without post-processing (third column), results with post-processing (fourth column).}
  \label{post}
\end{figure}

We show visual comparisons before and after post-processing in Figure~\ref{post}. The~results demonstrated that the ``hole'' (the first row in Figure~\ref{post}) was effectively filled, and~the ``island'' (the second row in Figure~\ref{post}) in the background was removed by post-processing. The~experiments indicated that our model-driven approach with post-processing was more effective. 

Qualitative and quantitative results demonstrated the effectiveness and advantages of the proposed method. Our method achieved a robust and accurate performance compared with the existing state-of-the-art methods in the 3D liver segmentation task, which can be applied to 3D medical image segmentation.

\section{Conclusion}\label{sect4}
In this work, we proposed a novel deep learning framework, FAS-UNet, for 2D and 3D medical image segmentation by enforcing some of the mathematical and physical laws (e.g., the~convexity Mumford--Shah model and FAS algorithm), which focuses on learning the multiscale image features to generate the segmentation results.

Compared with other existing works that analyzed the connection between the multigrid and CNN, FAS-UNet integrates medical image data and mathematical models and~enhances the connection between data-driven and traditional variational model methodologies; it provides a helpful viewpoint for designing image segmentation network architectures. Compared with UNet, the~proposed FAS-UNet introduces the concept of the data space, which exploits the model prior information to extract the features. Specifically, the feature extraction task leads to solving nonlinear equations, and an~iterative scheme of numerical algorithms was designed to learn the features.
Our experimental results showed that the proposed method is able to improve medical image segmentation in different tasks, including the segmentation of thoracic organs at risk, the whole-heart and great vessel, and~liver segmentation. It is believed that the approach is a general one and can be applied to other image processing tasks, such as image denoising and image reconstruction. 
In addition, we found that the topological interaction module proposed by~\cite{gupta2022learning} can effectively improve the performance of many segmentation methods. Therefore, we will use this module in FAS-UNet to improve its performance in the future work.

\bibliographystyle{IEEEtran}
\bibliography{FAS_Unet_refs}

\begin{thebibliography}{10}
\providecommand{\url}[1]{#1}
\csname url@samestyle\endcsname
\providecommand{\newblock}{\relax}
\providecommand{\bibinfo}[2]{#2}
\providecommand{\BIBentrySTDinterwordspacing}{\spaceskip=0pt\relax}
\providecommand{\BIBentryALTinterwordstretchfactor}{4}
\providecommand{\BIBentryALTinterwordspacing}{\spaceskip=\fontdimen2\font plus
\BIBentryALTinterwordstretchfactor\fontdimen3\font minus
  \fontdimen4\font\relax}
\providecommand{\BIBforeignlanguage}[2]{{%
\expandafter\ifx\csname l@#1\endcsname\relax
\typeout{** WARNING: IEEEtran.bst: No hyphenation pattern has been}%
\typeout{** loaded for the language `#1'. Using the pattern for}%
\typeout{** the default language instead.}%
\else
\language=\csname l@#1\endcsname
\fi
#2}}
\providecommand{\BIBdecl}{\relax}
\BIBdecl

\bibitem{minaee_ImageSegmentation_2021a}
\BIBentryALTinterwordspacing
S.~Minaee, Y.~Y. Boykov, F.~Porikli, A.~J. Plaza, N.~Kehtarnavaz, and
  D.~Terzopoulos, ``Image {{Segmentation Using Deep Learning}}: {{A Survey}},''
  pp. 1--1. [Online]. Available:
  \url{https://ieeexplore.ieee.org/document/9356353/}
\BIBentrySTDinterwordspacing

\bibitem{boveiri_MedicalImage_2020}
\BIBentryALTinterwordspacing
H.~R. Boveiri, R.~Khayami, R.~Javidan, and A.~Mehdizadeh, ``Medical image
  registration using deep neural networks: {{A}} comprehensive review,''
  vol.~87, p. 106767. [Online]. Available:
  \url{https://linkinghub.elsevier.com/retrieve/pii/S0045790620306224}
\BIBentrySTDinterwordspacing

\bibitem{cai_ReviewApplication_2020}
\BIBentryALTinterwordspacing
L.~Cai, J.~Gao, and D.~Zhao, ``A review of the application of deep learning in
  medical image classification and segmentation,'' vol.~8, no.~11, pp.
  713--713. [Online]. Available:
  \url{http://atm.amegroups.com/article/view/36944/html}
\BIBentrySTDinterwordspacing

\bibitem{chen_DeepLearning_2020}
\BIBentryALTinterwordspacing
C.~Chen, C.~Qin, H.~Qiu, G.~Tarroni, J.~Duan, W.~Bai, and D.~Rueckert, ``Deep
  learning for cardiac image segmentation: {{A}} review,'' vol.~7, p.~25.
  [Online]. Available: \url{http://arxiv.org/abs/1911.03723}
\BIBentrySTDinterwordspacing

\bibitem{litjens_SurveyDeep_2017}
\BIBentryALTinterwordspacing
G.~Litjens, T.~Kooi, B.~E. Bejnordi, A.~A.~A. Setio, F.~Ciompi, M.~Ghafoorian,
  J.~A. van~der Laak, B.~van Ginneken, and C.~I. Sánchez, ``A survey on deep
  learning in medical image analysis,'' vol.~42, pp. 60--88. [Online].
  Available:
  \url{https://linkinghub.elsevier.com/retrieve/pii/S1361841517301135}
\BIBentrySTDinterwordspacing

\bibitem{miotto_DeepLearning_2018}
\BIBentryALTinterwordspacing
R.~Miotto, F.~Wang, S.~Wang, X.~Jiang, and J.~T. Dudley, ``Deep learning for
  healthcare: Review, opportunities and challenges,'' vol.~19, no.~6, pp.
  1236--1246. [Online]. Available:
  \url{https://academic.oup.com/bib/article/19/6/1236/3800524}
\BIBentrySTDinterwordspacing

\bibitem{fu_DeepLearning_2020}
\BIBentryALTinterwordspacing
Y.~Fu, Y.~Lei, T.~Wang, W.~J. Curran, T.~Liu, and X.~Yang, ``Deep learning in
  medical image registration: A review,'' vol.~65, no.~20, p. 20TR01. [Online].
  Available: \url{https://iopscience.iop.org/article/10.1088/1361-6560/ab843e}
\BIBentrySTDinterwordspacing

\bibitem{liu_ReviewDeepLearningBased_2021}
\BIBentryALTinterwordspacing
X.~Liu, L.~Song, S.~Liu, and Y.~Zhang, ``A {{Review}} of {{Deep-Learning-Based
  Medical Image Segmentation Methods}},'' vol.~13, no.~3, p. 1224. [Online].
  Available: \url{https://www.mdpi.com/2071-1050/13/3/1224}
\BIBentrySTDinterwordspacing

\bibitem{UNet}
O.~Ronneberger, P.~Fischer, and T.~Brox, ``U-net: Convolutional networks for
  biomedical image segmentation,'' in \emph{MICCAI}, 2015.

\bibitem{milletari_VNetFully_2016a}
\BIBentryALTinterwordspacing
F.~Milletari, N.~Navab, and S.-A. Ahmadi, ``V-{{Net}}: {{Fully Convolutional
  Neural Networks}} for {{Volumetric Medical Image Segmentation}},'' in
  \emph{2016 {{Fourth International Conference}} on {{3D Vision}}
  ({{3DV}})}.\hskip 1em plus 0.5em minus 0.4em\relax {IEEE}, pp. 565--571.
  [Online]. Available: \url{http://ieeexplore.ieee.org/document/7785132/}
\BIBentrySTDinterwordspacing

\bibitem{zhou2018nest}
Z.~Zhou, M.~M.~R. Siddiquee, N.~Tajbakhsh, and J.~Liang, ``Unet++: Redesigning
  skip connections to exploit multiscale features in image segmentation,''
  \emph{IEEE transactions on medical imaging}, vol.~39, no.~6, pp. 1856--1867,
  2019.

\bibitem{cicek_3DUNet_2016a}
\BIBentryALTinterwordspacing
O.~Cicek, A.~Abdulkadir, S.~S. Lienkamp, T.~Brox, and O.~Ronneberger, ``{{3D
  U-Net}}: {{Learning Dense Volumetric Segmentation}} from {{Sparse
  Annotation}},'' in \emph{Medical {{Image Computing}} and {{Computer-Assisted
  Intervention}} - {{MICCAI}} 2016}, ser. Lecture {{Notes}} in {{Computer
  Science}}, S.~Ourselin, L.~Joskowicz, M.~R. Sabuncu, G.~Unal, and W.~Wells,
  Eds.\hskip 1em plus 0.5em minus 0.4em\relax {Springer International
  Publishing}, vol. 9901, pp. 424--432. [Online]. Available:
  \url{https://link.springer.com/10.1007/978-3-319-46723-8_49}
\BIBentrySTDinterwordspacing

\bibitem{SMehta2018}
S.~Mehta, E.~Mercan, J.~Bartlett, D.~Weaver, J.~G. Elmore, and L.~Shapiro,
  ``Y-net: {{Joint}} segmentation and classification for diagnosis of breast
  biopsy images,'' in \emph{Medical Image Computing and Computer Assisted
  Intervention – {{MICCAI}} 2018}, A.~F. Frangi, J.~A. Schnabel,
  C.~Davatzikos, C.~Alberola-López, and G.~Fichtinger, Eds.\hskip 1em plus
  0.5em minus 0.4em\relax {Springer International Publishing}, pp. 893--901.

\bibitem{xiao_WeightedResUNet_2018}
\BIBentryALTinterwordspacing
X.~Xiao, S.~Lian, Z.~Luo, and S.~Li, ``Weighted {{Res-UNet}} for {{High-Quality
  Retina Vessel Segmentation}},'' in \emph{2018 9th {{International
  Conference}} on {{Information Technology}} in {{Medicine}} and {{Education}}
  ({{ITME}})}.\hskip 1em plus 0.5em minus 0.4em\relax {IEEE}, pp. 327--331.
  [Online]. Available: \url{https://ieeexplore.ieee.org/document/8589312/}
\BIBentrySTDinterwordspacing

\bibitem{valanarasu_KiUNetAccurate_2020}
\BIBentryALTinterwordspacing
J.~M.~J. Valanarasu, V.~A. Sindagi, I.~Hacihaliloglu, and V.~M. Patel,
  ``{{KiU-Net}}: {{Towards Accurate Segmentation}} of {{Biomedical Images Using
  Over-Complete Representations}},'' in \emph{Medical {{Image Computing}} and
  {{Computer Assisted Intervention}} – {{MICCAI}} 2020}, ser. Lecture
  {{Notes}} in {{Computer Science}}, A.~L. Martel, P.~Abolmaesumi, D.~Stoyanov,
  D.~Mateus, M.~A. Zuluaga, S.~K. Zhou, D.~Racoceanu, and L.~Joskowicz,
  Eds.\hskip 1em plus 0.5em minus 0.4em\relax {Springer International
  Publishing}, vol. 12264, pp. 363--373. [Online]. Available:
  \url{https://link.springer.com/10.1007/978-3-030-59719-1_36}
\BIBentrySTDinterwordspacing

\bibitem{li_HDenseUNetHybrid_2018a}
\BIBentryALTinterwordspacing
X.~Li, H.~Chen, X.~Qi, Q.~Dou, C.-W. Fu, and P.-A. Heng, ``H-{{DenseUNet}}:
  {{Hybrid Densely Connected UNet}} for {{Liver}} and {{Tumor Segmentation From
  CT Volumes}},'' vol.~37, no.~12, pp. 2663--2674. [Online]. Available:
  \url{https://ieeexplore.ieee.org/document/8379359/}
\BIBentrySTDinterwordspacing

\bibitem{IsenseeF_2021}
\BIBentryALTinterwordspacing
F.~Isensee, P.~F. Jaeger, S.~A.~A. Kohl, J.~Petersen, and K.~H. Maier-Hein,
  ``{{nnU-Net}}: A self-configuring method for deep learning-based biomedical
  image segmentation,'' vol.~18, no.~2, pp. 203--211. [Online]. Available:
  \url{http://www.nature.com/articles/s41592-020-01008-z}
\BIBentrySTDinterwordspacing

\bibitem{He2019ODEInspiredND}
X.~He, Z.~Mo, P.~Wang, Y.~Liu, M.~Yang, and J.~Cheng, ``Ode-inspired network
  design for single image super-resolution,'' \emph{2019 IEEE/CVF Conference on
  Computer Vision and Pattern Recognition (CVPR)}, pp. 1732--1741, 2019.

\bibitem{pmlr-v80-lu18d}
\BIBentryALTinterwordspacing
Y.~Lu, A.~Zhong, Q.~Li, and B.~Dong, ``Beyond finite layer neural networks:
  Bridging deep architectures and numerical differential equations,'' ser.
  Proceedings of Machine Learning Research, J.~Dy and A.~Krause, Eds.,
  vol.~80.\hskip 1em plus 0.5em minus 0.4em\relax StockholmsmÃ¤ssan,
  Stockholm Sweden: PMLR, 10--15 Jul 2018, pp. 3276--3285. [Online]. Available:
  \url{http://proceedings.mlr.press/v80/lu18d.html}
\BIBentrySTDinterwordspacing

\bibitem{canet2020}
R.~Gu, G.~Wang, T.~Song, R.~Huang, M.~Aertsen, J.~Deprest, S.~Ourselin,
  T.~Vercauteren, and S.~Zhang, ``Ca-net: Comprehensive attention convolutional
  neural networks for explainable medical image segmentation,'' \emph{IEEE
  transactions on medical imaging}, vol.~40, no.~2, pp. 699--711, 2020.

\bibitem{he2016deep}
K.~He, X.~Zhang, S.~Ren, and J.~Sun, ``Deep residual learning for image
  recognition,'' in \emph{Proceedings of the IEEE conference on computer vision
  and pattern recognition}, 2016, pp. 770--778.

\bibitem{larsson2017fractalnet}
G.~Larsson, M.~Maire, and G.~Shakhnarovich, ``Fractalnet: Ultra-deep neural
  networks without residuals,'' in \emph{ICLR}, 2017.

\bibitem{zhang2017polynet}
X.~Zhang, Z.~Li, C.~Change~Loy, and D.~Lin, ``Polynet: A pursuit of structural
  diversity in very deep networks,'' in \emph{Proceedings of the IEEE
  Conference on Computer Vision and Pattern Recognition}, 2017, pp. 718--726.

\bibitem{gomez2017reversible}
A.~N. Gomez, M.~Ren, R.~Urtasun, and R.~B. Grosse, ``The reversible residual
  network: Backpropagation without storing activations,'' \emph{arXiv preprint
  arXiv:1707.04585}, 2017.

\bibitem{chen2018neural}
R.~T. Chen, Y.~Rubanova, J.~Bettencourt, and D.~Duvenaud, ``Neural ordinary
  differential equations,'' \emph{arXiv preprint arXiv:1806.07366}, 2018.

\bibitem{Yang0LX16}
Y.~Yang, J.~Sun, H.~Li, and Z.~Xu, ``Deep admm-net for compressive sensing
  {MRI},'' in \emph{Advances in Neural Information Processing Systems 29:
  Annual Conference on Neural Information Processing Systems 2016, December
  5-10, 2016, Barcelona, Spain}, D.~D. Lee, M.~Sugiyama, U.~von Luxburg,
  I.~Guyon, and R.~Garnett, Eds., 2016, pp. 10--18.

\bibitem{He2019MgNetAU}
J.~He and J.~Xu, ``Mgnet: A unified framework of multigrid and convolutional
  neural network,'' \emph{Science China Mathematics}, vol.~62, pp. 1331--1354,
  2019.

\bibitem{alt2021connections}
T.~Alt, K.~Schrader, M.~Augustin, P.~Peter, and J.~Weickert, ``Connections
  between numerical algorithms for pdes and neural networks,'' \emph{arXiv
  preprint arXiv:2107.14742}, 2021.

\bibitem{GHatfield2002}
G.~Hatfield, ``Perception as unconscious inference,'' in \emph{Perception and
  the physical world: Psychological and philosophical issues in
  perception}.\hskip 1em plus 0.5em minus 0.4em\relax Citeseer, 2002.

\bibitem{XHCai2013}
\BIBentryALTinterwordspacing
X.~Cai, R.~Chan, and T.~Zeng, ``A two-stage image segmentation method using a
  convex variant of the mumford--shah model and thresholding,'' \emph{SIAM
  Journal on Imaging Sciences}, vol.~6, no.~1, pp. 368--390, 2013. [Online].
  Available: \url{https://doi.org/10.1137/120867068}
\BIBentrySTDinterwordspacing

\bibitem{liu_WeightedVariational_2018}
\BIBentryALTinterwordspacing
C.~Liu, M.~K.-P. Ng, and T.~Zeng, ``Weighted variational model for selective
  image segmentation with application to medical images,'' vol.~76, pp.
  367--379. [Online]. Available:
  \url{https://linkinghub.elsevier.com/retrieve/pii/S0031320317304715}
\BIBentrySTDinterwordspacing

\bibitem{ma_ImageSegmentation_2018}
\BIBentryALTinterwordspacing
Q.~Ma, J.~Peng, and D.~Kong, ``Image {{Segmentation}} via {{Mean Curvature
  Regularized Mumford-Shah Model}} and {{Thresholding}},'' vol.~48, no.~2, pp.
  1227--1241. [Online]. Available:
  \url{http://link.springer.com/10.1007/s11063-017-9763-7}
\BIBentrySTDinterwordspacing

\bibitem{mccormick1987multigrid}
S.~F. McCormick, \emph{Multigrid methods}.\hskip 1em plus 0.5em minus
  0.4em\relax SIAM, 1987.

\bibitem{mumford_OptimalApproximations_1989a}
D.~B. Mumford and J.~Shah, ``Optimal approximations by piecewise smooth
  functions and associated variational problems,'' \emph{Communications on pure
  and applied mathematics}, 1989.

\bibitem{ZHOU2020787}
\BIBentryALTinterwordspacing
D.-X. Zhou, ``Universality of deep convolutional neural networks,''
  \emph{Applied and Computational Harmonic Analysis}, vol.~48, no.~2, pp.
  787--794, 2020. [Online]. Available:
  \url{https://www.sciencedirect.com/science/article/pii/S1063520318302045}
\BIBentrySTDinterwordspacing

\bibitem{taha2015metrics}
A.~A. Taha and A.~Hanbury, ``Metrics for evaluating 3d medical image
  segmentation: analysis, selection, and tool,'' \emph{BMC Medical Imaging},
  vol.~15, no.~1, pp. 29--29, 2015.

\bibitem{Dice1945Measures}
L.~R. Dice, ``Measures of the amount of ecologic association between species,''
  \emph{Ecology}, vol.~26, no.~3, 1945.

\bibitem{segthor}
Z.~Lambert, C.~Petitjean, B.~Dubray, and S.~Kuan, ``Segthor: Segmentation of
  thoracic organs at risk in ct images,'' in \emph{2020 Tenth International
  Conference on Image Processing Theory, Tools and Applications (IPTA)}, 2020,
  pp. 1--6.

\bibitem{pace2015interactive}
D.~F. Pace, A.~V. Dalca, T.~Geva, A.~J. Powell, M.~H. Moghari, and P.~Golland,
  ``Interactive whole-heart segmentation in congenital heart disease,'' vol.
  9351, pp. 80--88, 2015.

\bibitem{kavur2020chaos}
A.~E. Kavur, N.~S. Gezer, M.~Bar{\i}{\c{s}}, S.~Aslan, P.-H. Conze, V.~Groza,
  D.~D. Pham, S.~Chatterjee, P.~Ernst, S.~{\"O}zkan \emph{et~al.}, ``Chaos
  challenge-combined (ct-mr) healthy abdominal organ segmentation,''
  \emph{Medical Image Analysis}, vol.~69, p. 101950, 2021.

\bibitem{CHAOSdata2019}
\BIBentryALTinterwordspacing
A.~E. Kavur, M.~A. Selver, O.~Dicle, M.~Barış, and N.~S. Gezer, ``{CHAOS -
  Combined (CT-MR) Healthy Abdominal Organ Segmentation Challenge Data},'' Apr.
  2019. [Online]. Available: \url{https://doi.org/10.5281/zenodo.3362844}
\BIBentrySTDinterwordspacing

\bibitem{gu2019ce-net}
Z.~Gu, J.~Cheng, H.~Fu, K.~Zhou, H.~Hao, Y.~Zhao, T.~Zhang, S.~Gao, and J.~Liu,
  ``Ce-net: Context encoder network for 2d medical image segmentation,''
  \emph{IEEE Transactions on Medical Imaging}, vol.~38, no.~10, pp. 2281--2292,
  2019.

\bibitem{cpfnet}
S.~Feng, H.~Zhao, F.~Shi, X.~Cheng, M.~Wang, Y.~Ma, D.~Xiang, W.~Zhu, and
  X.~Chen, ``Cpfnet: Context pyramid fusion network for medical image
  segmentation,'' \emph{IEEE Transactions on Medical Imaging}, vol.~39, no.~10,
  pp. 3008--3018, 2020.

\bibitem{romera2018erfnet}
E.~Romera, J.~M. Alvarez, L.~M. Bergasa, and R.~Arroyo, ``Erfnet: Efficient
  residual factorized convnet for real-time semantic segmentation,'' \emph{IEEE
  Transactions on Intelligent Transportation Systems}, vol.~19, no.~1, pp.
  263--272, 2018.

\bibitem{linknet}
A.~Chaurasia and E.~Culurciello, ``Linknet: Exploiting encoder representations
  for efficient semantic segmentation,'' in \emph{2017 IEEE Visual
  Communications and Image Processing (VCIP)}.\hskip 1em plus 0.5em minus
  0.4em\relax IEEE, 2017, pp. 1--4.

\bibitem{3DUnet}
O.~Cicek, A.~Abdulkadir, S.~S. Lienkamp, T.~Brox, and O.~Ronneberger, ``3d
  u-net: Learning dense volumetric segmentation from sparse annotation,'' pp.
  424--432, 2016.

\bibitem{2019AnatomyNet}
W.~Zhu, Y.~Huang, L.~Zeng, X.~Chen, Y.~Liu, Z.~Qian, N.~Du, W.~Fan, and X.~Xie,
  ``Anatomynet: Deep learning for fast and fully automated whole‐volume
  segmentation of head and neck anatomy,'' \emph{Medical Physics}, vol.~46,
  2019.

\bibitem{chen2019dmfnet}
C.~Chen, X.~Liu, M.~Ding, J.~Zheng, and J.~Li, ``3d dilated multi-fiber network
  for real-time brain tumor segmentation in mri,'' in \emph{International
  Conference on Medical Image Computing and Computer Assisted Intervention
  (MICCAI)}, 2019.

\bibitem{luo2020hdc}
Z.~Luo, Z.~Jia, Z.~Yuan, and J.~Peng, ``Hdc-net: Hierarchical decoupled
  convolution network for brain tumor segmentation,'' \emph{IEEE Journal of
  Biomedical and Health Informatics}, vol.~25, no.~3, pp. 737--745, 2020.

\bibitem{zhang2019rsanet}
H.~Zhang, J.~Zhang, Q.~Zhang, J.~Kim, S.~Zhang, S.~A. Gauthier,
  P.~Spincemaille, T.~D. Nguyen, M.~Sabuncu, and Y.~Wang, ``Rsanet: Recurrent
  slice-wise attention network for multiple sclerosis lesion segmentation,'' in
  \emph{International Conference on Medical Image Computing and
  Computer-Assisted Intervention}.\hskip 1em plus 0.5em minus 0.4em\relax
  Springer, 2019, pp. 411--419.

\bibitem{bui2019skip-connected}
T.~D. Bui, J.~Shin, and T.~Moon, ``Skip-connected 3d densenet for volumetric
  infant brain mri segmentation,'' \emph{Biomedical Signal Processing and
  Control}, vol.~54, p. 101613, 2019.

\bibitem{Chen2018VoxResNetDV}
H.~Chen, Q.~Dou, L.~Yu, J.~Qin, and P.-A. Heng, ``Voxresnet: Deep voxelwise
  residual networks for brain segmentation from 3d mr images,''
  \emph{NeuroImage}, vol. 170, pp. 446--455, 2018.

\bibitem{mehta20193despnet}
N.~Nuechterlein and S.~Mehta, ``3d-espnet with pyramidal refinement for
  volumetric brain tumor image segmentation,'' in \emph{International MICCAI
  Brainlesion Workshop}.\hskip 1em plus 0.5em minus 0.4em\relax Springer, 2018,
  pp. 245--253.

\bibitem{valverde2020ratlesnetv2}
J.~M. Valverde, A.~Shatillo, R.~De~Feo, O.~Gr{\"o}hn, A.~Sierra, and J.~Tohka,
  ``Ratlesnetv2: a fully convolutional network for rodent brain lesion
  segmentation,'' \emph{Frontiers in neuroscience}, p. 1333, 2020.

\bibitem{gupta2022learning}
S.~Gupta, X.~Hu, J.~Kaan, M.~Jin, M.~Mpoy, K.~Chung, G.~Singh, M.~Saltz,
  T.~Kurc, J.~Saltz \emph{et~al.}, ``Learning topological interactions for
  multi-class medical image segmentation,'' \emph{arXiv preprint
  arXiv:2207.09654}, 2022.

\end{thebibliography}

\end{document}